\documentclass[preprint,12pt]{elsarticle}

\usepackage{amssymb}
\usepackage{amsmath}
\journal{} 

\usepackage{lineno}

\usepackage{array}
\newcolumntype{C}{>{\centering\arraybackslash}X}

\newlength{\extralength}  % Define \extralength
\newlength{\fulllength}  % Define \fulllength
\usepackage{makecell}

\usepackage[super]{nth}
\usepackage{amsmath}
\usepackage{algorithm}
\usepackage{algpseudocode}
\usepackage{graphicx} 
\usepackage{subcaption}
\usepackage{longtable}
\usepackage{booktabs}
\usepackage{rotating}
\usepackage{amssymb} % For \mathfrak
\usepackage{tabularx}

\usepackage{tabularx}
\usepackage{adjustbox} % For adjustwidth
\usepackage{pdflscape} % For landscape orientation
\usepackage{multirow}   % If using multi-row cells
\usepackage{adjustbox}
\usepackage{tabularx}
\newcolumntype{Y}{>{\centering\arraybackslash}X}

\usepackage{booktabs}
\usepackage{changepage} % For adjustwidth
\usepackage{graphicx}   % If you include images

\begin{document}

\makeatletter
\def\ps@pprintTitle{%
 \let\@oddhead\@empty
 \let\@evenhead\@empty
 \let\@oddfoot\@empty
 \let\@evenfoot\@empty}
\makeatother

\begin{frontmatter}

\ExplSyntaxOn
\cs_gset:Npn \__first_footerline:
  { \group_begin: \small \sffamily \__short_authors: \group_end: }
\ExplSyntaxOff

%% Title, authors and addresses

%% use the tnoteref command within \title for footnotes;
%% use the tnotetext command for theassociated footnote;
%% use the fnref command within \author or \affiliation for footnotes;
%% use the fntext command for theassociated footnote;
%% use the corref command within \author for corresponding author footnotes;
%% use the cortext command for theassociated footnote;
%% use the ead command for the email address,
%% and the form \ead[url] for the home page:
%% \title{Title\tnoteref{label1}}
%% \tnotetext[label1]{}
%% \author{Name\corref{cor1}\fnref{label2}}
%% \ead{email address}
%% \ead[url]{home page}
%% \fntext[label2]{}
%% \cortext[cor1]{}
%% \affiliation{organization={},
%%             addressline={},
%%             city={},
%%             postcode={},
%%             state={},
%%             country={}}
%% \fntext[label3]{}

%\title{Tab2Visual: Classification Via Visual Representation Of Tabular Data} %% Article title
\title{Tab2Visual: Overcoming Limited Data in Tabular Data Classification Using Deep Learning with Visual Representations}
%% use optional labels to link authors explicitly to addresses:
%% \author[label1,label2]{}
%% \affiliation[label1]{organization={},
%%             addressline={},
%%             city={},
%%             postcode={},
%%             state={},
%%             country={}}
%%
%% \affiliation[label2]{organization={},
%%             addressline={},
%%             city={},
%%             postcode={},
%%             state={},
%%             country={}}

\author[1]{Ahmed Mamdouh}
\ead{ahmed.mamdouh@aun.edu.eg}
\author[1,2]{Moumen El-Melegy}
\ead{melmelegy@bwh.harvard.edu}
\author[1]{Samia Ali}
\ead{samia_fattah@aun.edu.eg}
\author[2]{Ron Kikinis}
\ead{kikinis@bwh.harvard.edu}
%\ead{ahmed.mamdouh@aun.edu.eg; melmelegy@bwh.harvard.edu; samia_fattah@aun.edu.eg; kikinis@bwh.harvard.edu}

%% Author affiliation
\affiliation[1]{organization={Electrical Engineering Department, Assiut University}, % Department and Organization
            city={Assiut},
            postcode={71516}, 
            %state={Assiut},
            country={Egypt}}
            
\affiliation[2]{organization={Brigham and Women’s Hospital, Harvard Medical School}, % Department and Organization
            city={Boston},
            state={MA},
            postcode={02115}, 
            country={USA}}

%% Abstract
\begin{abstract}
  %% Text of abstract
  This research addresses the challenge of limited data in tabular data classification, particularly prevalent in domains with constraints like healthcare. We propose Tab2Visual, a novel approach that transforms heterogeneous tabular data into visual representations, enabling the application of powerful deep learning models. Tab2Visual effectively addresses data scarcity by incorporating novel image augmentation techniques and facilitating transfer learning. 
  We extensively evaluate the proposed approach on diverse tabular datasets, comparing its performance against a wide range of machine learning algorithms, including classical methods, tree-based ensembles, and state-of-the-art deep learning models specifically designed for tabular data. 
 We also perform an in-depth analysis of factors influencing Tab2Visual's performance.  
 Our experimental results demonstrate that Tab2Visual outperforms other methods in classification problems with limited tabular data.
 %Our experimental results demonstrate the effectiveness of Tab2Visual over the other methods in scenarios with limited data. 
%
%
  %We introduce Tab2Visual, a novel method that converts tabular data into image representations, harnessing the power of advanced computer vision techniques, including convolutional neural networks (CNNs) and Vision Transformers (ViTs). This approach leverages image augmentation strategies to expand the dataset and enhance the model's ability to generalize, particularly in scenarios involving small datasets. Our experiments show that Tab2Visual outperforms or matches the performance of traditional tree-based methods and state-of-the-art deep learning techniques when applied to smaller datasets, highlighting its effectiveness in these contexts. However, its performance varies with larger datasets, where the benefits of augmentation are diminished, and the complexity of the data favors more specialized approaches. Despite this, Tab2Visual offers a significant step forward in the application of deep learning models to tabular data, providing a versatile and powerful tool for improving prediction accuracy across a range of dataset sizes.
\end{abstract}

%%%%%%%%%%%%%%%%%%%%%%%%%%%%%%%%%%%%%%%%%%%%%%%%%%%%%%%%%%%%%%%%5
%%Graphical abstract
%\begin{graphicalabstract}
%\includegraphics[width=1.0\columnwidth]{tab2visual_vis_abstract.png}
%\end{graphicalabstract}

%%%%%%%%%%%%%%%%%%%%%%%%%%%%%%%%%%%%%%%%%%%%%%%%%%%%%%%%%%%%%%5
%%Research highlights
%\begin{highlights}
%\item Tab2Visual transforms heterogeneous tabular data into images for effective deep learning classification. 
%\item Tab2Visual enables enhanced data augmentation and efficient transfer learning.
%\item A set of efficient and meaningful image augmentation techniques is designed.
%\item Rigorous evaluation is performed against other machine learning algorithms on diverse datasets.
%\end{highlights}

%% Keywords
\begin{keyword}
  %% keywords here, in the form: keyword \sep keyword
  Tabular Data, Limited Data, Deep Learning, Machine Learning, Data Augmentation, Transfer Learning 
  
  %% PACS codes here, in the form: \PACS code \sep code
  %% MSC codes here, in the form: \MSC code \sep code
  %\MSC[2020] 68T05 \sep 68T10
  \end{keyword}
  
\end{frontmatter}

%% Add \usepackage{lineno} before \begin{document} and uncomment 
%% following line to enable line numbers
%% \linenumbers

%% main text
%%

%%%%%%%%%%%%%%%%%%%%%%%%%%%%%%%%%%%%%%%%%%%%%%%%%%%%%%%%%%%%%%%%%%%%%%%%%%%%%%%%%%%%%%%%%%%%%%%%%%%%%%%%%%%%%%%%%%%%%%%%%%%%%%%
%% main text
\section{Introduction}

Recent years have witnessed significant advancements in deep learning, with Convolutional Neural Networks (CNNs) and Vision Transformers (ViTs) revolutionizing image classification tasks~\cite{krizhevsky2012imagenet}.
CNNs excel at automatically extracting hierarchical features from raw data, thus being exceptionally effective in handling the complexities of image data~\cite{he2016deep}. 
ViTs have further pushed the boundaries by capturing long-range dependencies and surpassing CNNs in several benchmarks, demonstrating their powerful capabilities~\cite{dosovitskiy2020image, carion2020end}. Both CNNs and ViTs have exceeded human performance in numerous tasks~\cite{silver2017mastering}, solidifying their status as state-of-the-art techniques in the field. 

However, deep learning has not been that successful on tabular data~\cite{Grinsztajn_NIPS2022, Borisov_NNLS2024}. 
Tabular data, commonly represented as a two-dimensional matrix of rows (samples or observations) and columns (attributes or features), constitute the most prevalent data format in data science~\cite{Breugel_ICML2024}.
%Tabular data, typically represented as a two-dimensionally input of rows (samples) and columns (features), are the most ubiquitous form of data in data sciences~\cite{Breugel_ICML2024}. 
Classification of these data has broad applications across finance, healthcare, cybersecurity, anomaly detection, and social science~\cite{Borisov_NNLS2024}.
The inherent characteristics of tabular data pose unique challenges for machine learning algorithms.
Unlike image, language, and speech data, which often exhibit inherent structure and homogeneity, tabular data typically comprises diverse feature types (numerical, categorical, Boolean, and ordinal) with varying scales and sources, leading to increased complexity.  Furthermore, the interdependencies among features in tabular data are generally weaker compared to those observed in image or speech data, where spatial or semantic relationships often establish strong correlations between data points.

Due to these challenges, deep neural networks frequently underperform on tabular data compared to other machine learning methods, such as tree-based ensembles.
To address these challenges, researchers have explored various approaches, broadly categorized into three groups~\cite{ Borisov_NNLS2024}: specialized architectures, regularization techniques, and data transformation methods.

The first group focuses on developing specialized architectures specifically designed for tabular data. Notable examples within this category include NODE~\cite{Popov2020Neural}, TabNet~\cite{arik2021tabnet}, TabTransformer~\cite{huang2020tabtransformer}, and TabPFN~\cite{hollmann2022tabpfn,Hollmann_Nature2025}. 
The Neural Oblivious Decision Ensembles (NODE)~\cite{Popov2020Neural} are characterized by a deep, layer-wise structure composed of an ensemble of differentiable oblivious trees. Gradient-based optimization is employed for end-to-end training of this architecture.
%The Neural Oblivious Decision Ensembles (NODE) architecture~\cite{Popov2020Neural} comprises an ensemble of differentiable oblivious decision trees organized in a deep, layer-wise structure. The ensemble is trained end-to-end using gradient-based optimization techniques.
The other three architectures, TabNet, TabTransformer, and TabPFN, are inspired by the success of transformer models. 
%Transformers excel at processing sequential data by leveraging attention mechanisms, enabling them to effectively capture long-range dependencies and intricate relationships within the sequence. 
%TabNet~\cite{arik2021tabnet} is one of the earliest transformer-based architectures for tabular data, utilizing a sequential attention mechanism.
TabNet~\cite{arik2021tabnet} is a pioneering transformer-based architecture for tabular data, employing a  sequential attention mechanism.
Similarly, the TabTransformer~\cite{huang2020tabtransformer} employs self-attention mechanisms within a transformer architecture to generate contextual embeddings for categorical features. These enriched embeddings, along with the original numerical features, are then fed into a multilayer perceptron for the final classification.
The more recent Tabular Prior-data Fitted Network (TabPFN)~\cite{hollmann2022tabpfn, Hollmann_Nature2025} relies on in-context learning that integrates approximate Bayesian inference and structural causal modeling within a two-way attention mechanism. Notably, TabPFN undergoes a single pre-training phase on a vast collection of synthetic datasets encompassing diverse prediction tasks. During inference, given a new dataset with both labeled training and unlabeled test samples, the model simultaneously trains and predicts within a single forward pass of the neural network.
Due to its significant memory requirements, TabPFN's current application is restricted to small datasets~\cite{Hollmann_Nature2025}.

The second group of tabular data methods attributes the moderate performance of deep learning models to their inherent nonlinearity and high model complexity. As such, they enforce strong regularization during model training. This often involves employing specialized loss functions, such as the learned regularization scheme proposed by Shavitt et al.~\cite{ Shavitt_NIPS2018} and the regularization cocktails (combinations) introduced by Kadra et al.~\cite{ Kadra_NIPS2021}.

The final category focuses on data transformation methods, aiming to convert heterogeneous tabular inputs into a homogeneous format more suitable for deep learning. These methods typically focus on data preprocessing techniques rather than requiring the development of entirely new deep architectures. One pioneering approach is DeepInsight~\cite{sharma2019deepinsight}, which transforms high-dimensional tabular data into a spatial representation suitable for CNNs. DeepInsight projects the data into a 2D space using t-SNE~\cite{JMLR:v9:vandermaaten08a}, a non-linear dimensionality reduction technique that preserves local similarities. DeepInsight then constructs an image from this 2D projection using convex hull analysis with proper translation, rotation, quantization, and normalization operations. This method relies on t-SNE, which is non-deterministic and sensitive to parameter choices (e.g., perplexity~\cite{JMLR:v9:vandermaaten08a}) and suffers from "crowding problem" where many points may cluster together in the low-dimensional space. 
%It is computationally expensive for large datasets due to pairwise distance calculations and the optimization process. 
Furthermore, for datasets with few features, the resulting images may contain isolated islands, reducing the efficiency of CNNs in learning meaningful patterns.

The REFINED (REpresentation of Features as Images with NEighborhood Dependencies) approach~\cite{Bazgir_NatComm20} also projects data into a 2D space, but replaces t-SNE with Bayesian Metric Multidimensional Scaling to preserve pairwise distances in the low-dimensional representation. Despite these modifications, REFINED shares similar limitations to DeepInsight.
The SuperTML method~\cite{Sun_SuperTML2019} adopts another approach by converting tabular data into text representations visualized as 2D binary images. This transforms the tabular classification problem into a text classification task, which can then be addressed by CNNs. However, the reported method's evaluation is limited to only three datasets.
	
Buturovic et al.~\cite{buturovic2020novel} introduced the TAbular Convolution (TAC) method, which arranges data samples into zero-mean square matrices (kernels) of odd integer dimensions. These kernels are then convolved with a fixed "base image," and the resulting images are subsequently fed to a CNN for classification. While applied successfully to classify gene expression data in their study, the method's reliance on an arbitrarily chosen base image presents a significant limitation. The impact of this base image on model performance remains unclear, and the authors themselves acknowledge its inconclusive influence. Furthermore, the necessary padding or trimming of data samples to achieve the required kernel size can potentially introduce bias or information loss, negatively impacting the method's overall performance.
Zhu et al.~\cite{nguyen2021image} proposed the Image Generator for Tabular Data (IGTD) approach. IGTD generates an image for each data sample, where pixel intensities directly represent feature values. 
Consequently, the resulting image has a size that corresponds to the number of features in the original data.
The algorithm employs an iterative optimization process to assign features to pixels, prioritizing the placement of similar features in close proximity. This approach leads to more compact image representations, reducing memory consumption and accelerating CNN training. However, IGTD faces limitations when applied to datasets with a limited number of features. In such cases, the generated images may lack sufficient detail, hindering effective CNN training.

Despite all those research efforts to apply deep learning to tabular data, a recent comprehensive review~\cite{Borisov_NNLS2024} concludes that tree-ensemble models (e.g., XGBoost, LightGBM, CatBoost) continue to demonstrate superior performance in classification tasks.
This observation has been corroborated by numerous independent studies (e.g., see \cite{Grinsztajn_NIPS2022, Shwartz_IF2022}). The underperformance of deep learning is particularly pronounced on small tabular datasets~\cite{Borisov_NNLS2024, Shwartz_IF2022}. 
In other areas of machine learning, such as image classification and natural language processing,
these challenges are effectively addressed through techniques like data augmentation~\cite{Xu_PR2023} to increase the dataset size and/or transfer learning~\cite{Tan_ICANN2018} to leverage knowledge learned from data-rich domains.
While few data augmentation techniques have been proposed for tabular data—such as SMOTE variants 
(with SMOTE-NC being the only variant that can handle discrete features)~\cite{Chawla_JAIR2002} and the latent space interpolation method by Darabi and Elor~\cite{Darabi_ArXiv2021}—their primary focus remains minority class oversampling through linear syntheses of existing samples. These methods, however, often yield only marginal gains in classification performance~\cite{Darabi_ArXiv2021}, underscoring a critical gap in the field. Consequently, the development of effective augmentation strategies tailored to tabular data remains an open research challenge~\cite{Borisov_NNLS2024}. Furthermore, despite its success in other domains, transfer learning has proven challenging to effectively apply to tabular data~\cite{Borisov_NNLS2024}.

%%In addition, effective transfer learning strategies in the context of tabular data is still an area of active research~\cite{Borisov_NNLS2024}.

\begin{figure}[!t]
        \centerline{\includegraphics[width=0.7\columnwidth]{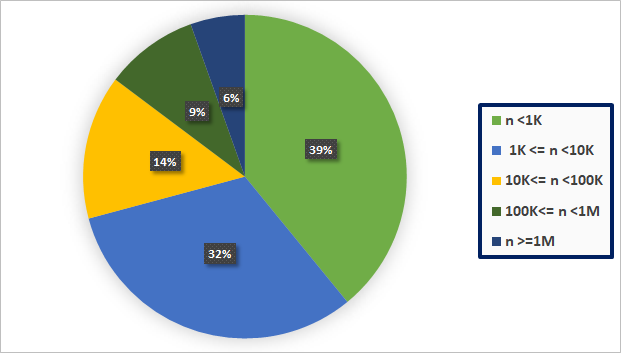}}
        \caption{Distribution of tabular dataset sizes on the OpenML platform ($n$ denotes the dataset size). The majority of datasets fall within the smaller size ranges, highlighting the significance of addressing challenges associated with limited data.} 
        \label{fig:openml statistics}
    \end{figure}

The current research is driven by the challenges of limited data encountered in our ongoing investigation of prostate cancer prediction using clinical biomarkers and patient-supplied questionnaires~\cite{mamdouh2022prediction, el2022prostate, el2024prostate}. Our dataset, with fewer than 100 records and only 9 features, exemplifies the typical data scarcity encountered in healthcare applications, often constrained by factors such as cost, privacy concerns, and limited access to resources. Our motivation is further reinforced by the prevalence of small tabular datasets across various domains. Notably, OpenML, a prominent platform for sharing machine learning datasets, reveals that over 71\% of available tabular datasets contain fewer than 10K records at the time of writing (see Fig.~\ref{fig:openml statistics}).

%%% Figure
    
To overcome these challenges, we introduce \emph{Tab2Visual}, a novel data transformation strategy that converts heterogeneous tabular data into visual representations, thereby enabling the utilization of powerful deep learning models including CNNs and ViTs.
This approach enables effective capture of complex patterns without specialized architectures. Furthermore, Tab2Visual offers potential solutions to open research questions regarding transfer learning and data augmentation for tabular data. Specifically, it leverages image augmentation to enhance model generalization and effectively increase dataset size without additional data collection. We also introduce a new set of efficient and semantically meaningful image augmentation techniques tailored for Tab2Visual-generated images. Additionally, Tab2Visual facilitates transfer learning, enabling fine-tuning of pre-trained models, even from different domains, for new tabular classification tasks. This reduces reliance on large labeled datasets and allows for efficient knowledge transfer between tabular datasets. Critically, unlike the aforementioned tabular-to-image conversion methods~\cite{sharma2019deepinsight, Bazgir_NatComm20, Sun_SuperTML2019, buturovic2020novel, nguyen2021image}, which typically require feature-rich datasets and struggle with generating meaningful augmentations, Tab2Visual effectively handles data with limited features and offers numerous augmentation strategies. Consequently, Tab2Visual effectively mitigates the challenges posed by limited data in tabular data classification.

It is interesting to note that the concept of representing tabular information as images, upon which Tab2Visual is founded, aligns with established principles in cognitive psychology. The Picture Superiority Effect~\cite{Paivio_1973, Mintzer_1999} demonstrates that people tend to learn and remember information presented visually more effectively than information presented in textual or numerical formats.
This phenomenon is explained by the fact that images trigger a richer set of cognitive representations and associations compared to words, drawing upon broader knowledge of the world~\cite{Grady_1998}. Furthermore, the distinctive visual features within images contribute to enhanced memorability~\cite{Mintzer_1999}.
%
%This phenomenon is attributed to the richer set of representations and associations triggered with other knowledge about the world compared to words~\cite{Grady_1998}. Furthermore, the distinctive visual features within images contribute to enhanced memorability~\cite{Mintzer_1999}.

This work significantly extends the preliminary, exploratory ideas drafted in our prior work~\cite{el2024prostate}, which focused on a specific clinical context. We make several new contributions, advancing the state-of-the-art across scope, algorithmic development, presentation, and experimental work. We address the wider context of general tabular data classification, tackling its open and challenging research problems.
In addition, we propose comprehensive and generalized algorithms for Tab2Visual modeling and image augmentation.
To demonstrate the effectiveness of Tab2Visual, we evaluate its performance using two state-of-the-art backbones: the CNN architecture, EfficientNet~\cite{tan2019efficientnet}, and the Vision Transformer, EfficientViT~\cite{liu2023efficientvit}. We conduct a comprehensive empirical study across 10 diverse datasets from the UCI Machine Learning Repository, spanning various application domains including medical, e-commerce, and engineering. While Tab2Visual is particularly intended for small datasets, we assess its performance on a range of datasets, from small to medium and large scale, encompassing thousands of records and tens of features.
Furthermore, we conduct a rigorous comparative analysis of Tab2Visual in terms of accuracy and speed against a diverse set of machine learning algorithms, including classical methods (support vector machines, logistic regression, shallow neural networks), established tree-based ensembles (random forests, extra trees, gradient boosting machines), and state-of-the-art deep learning architectures specifically designed for tabular data, such as TabNet~\cite{arik2021tabnet} and TabPFN~\cite{hollmann2022tabpfn, Hollmann_Nature2025}. This comprehensive evaluation aims to benchmark Tab2Visual against key existing approaches in the context of tabular data classification.
Finally, we perform an in-depth analysis of key factors influencing Tab2Visual's performance, including the impact of augmentation strategies, the choice between transfer learning and training from scratch, the selection of the backbone model, and the arrangement of features within the generated image representations.

The rest of this paper is structured as follows: Section~\ref{sec:t2v} introduces the Tab2Visual modeling approach. Section~\ref{sec:experiments} presents our comprehensive experimental setup and results, followed by thorough discussion and analysis in Section~\ref{sec:discussion}. 
Lastly, Section~\ref{sec:conclusions} provides a summary of our findings and concluding remarks.

%%%%%%%%%%%%%%%%%%%%%%%%%%%%%%%%%%%%%%%%%%%%%%%%%%%%%%%%%%%%%%%%%%%%%%%%%%%%%%%%%%%%%%%
\section{Tab2Visual: Visual Representation of Tabular Data}
\label{sec:t2v}

%Deep learning models have demonstrated remarkable success in image analysis by effectively leveraging spatial structure and local correlations within image data. 
Deep learning has achieved remarkable success in image analysis by effectively exploiting the spatial structure and local correlations inherent within image data.
However, the application of deep learning to heterogeneous tabular data has been less successful. Tab2Visual addresses this limitation by transforming tabular data into visual representations. This transformation enables powerful vision models, such as CNNs and ViTs, to effectively extract meaningful features from the tabular data, leading to improved classification performance.
This section details the proposed Tab2Visual methodology. We begin by formally defining the problem.

\textbf{Problem Statement:} Given a tabular dataset \( \mathcal{D} = \{ (\textbf{x}_i, y_i) \}_{i=1}^n \), where \( n \) is the number of samples, each sample \( \textbf{x}_i = [ x_i^{(1)}, x_i^{(2)}, \dots, x_i^{(m)}]^{T}\in \mathbb{R}^m \) consists of \( m \) features, and \( y_i \in \{1, \dots, C\} \) is the label associated with sample \( \textbf{x}_i \), where $C$ is the number of unique labels or classes,
the objective is to transform each sample \( \textbf{x}_i \) into an image representation, \( I_i \) of dimensions \( H \times W \), with \( H \) being the image height and \( W \) the image width, such that a deep learning model \( \mathcal{M} \) (e.g., CNN or ViT) can be trained on the set of images \( \{ (I_i, y_i) \}_{i=1}^n \) for the classification task.
%
%image dimensions \( h \times w \), where \( h \) is the image height and \( w \) is the image width; and a specified number of rows \( r \) for arranging features in the image representation, 

Fig.~\ref{fig:galaxy42} illustrates the Tab2Visual modeling steps, which are further elaborated below:
%The detailed steps for the proposed Tab2Visual modeling are illustrated in Fig.~\ref{fig:galaxy42} and explained as follows:

\begin{figure}[!t]
    \centerline{\includegraphics[width=\columnwidth]{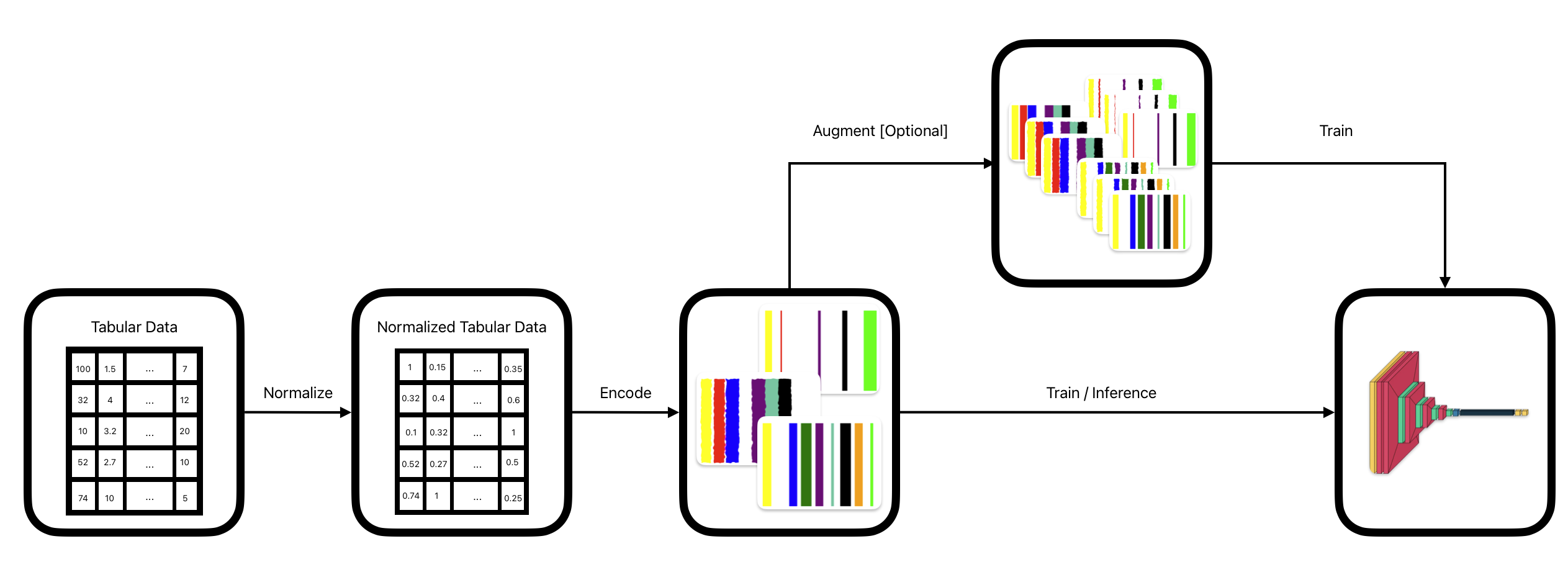}}
    \caption{The Tab2Visual approach: Tabular data undergoes normalization and is subsequently encoded as images where each feature is represented by a bar with varying width. To enhance data diversity, optional data augmentation techniques can be applied. Finally, these image representations are fed to a deep learning model for training or inference.}
    %Tab2Visual approach: Tabular data are normalized then encoded as images of width-varying bars. These images are optionally augmented to increase the data size before training. Images are then fed to a deep model for training or inference.} %TODO: Modify figure to account for inference pipeline as well.}
    \label{fig:galaxy42}
\end{figure}

\begin{itemize}
    \item \textbf{Data Preparation and Normalization:} 
    First, categorical features are one-hot encoded, and ordinal variables are encoded numerically based on their inherent order.
    Following this, min-max normalization is applied to the entire dataset \( \mathcal{D} \), such that
    \( x_i^{(j)} \in [0, 1], \forall i=1,...,n, j=1,...,m \). This standardization ensures consistent feature comparisons and prepares the data for visual representation.
    
    \item \textbf{Image Preparation:} The next step involves converting the normalized data into visual representations suitable for deep learning models. For each sample, an image is created encoding the sample's features. The image size is a user-supplied parameter, and should balance the desired image details with computational efficiency. It should also be compatible with the input requirements of the deep models to be used. \\
    \indent Generally, the image will consist of bars of widths proportional to the feature values, arranged in multiple rows and columns. The user specifies the desired number of rows, \( r \). Accordingly, the number of bars per row becomes \( c = \lceil m / r \rceil \). By construction, the image space is divided equally among all features; the maximum bar width is calculated as \( b = W / c \), and each bar has a height of  \( h = H / r \). 
    For example, when $r=1$, a dataset with 9 features will be represented as an image with 9 bars arranged in a single row, as illustrated in Fig.~\ref{fig:galaxy9}.
    For datasets with a larger number of features, to ensure the bars have sufficient image support, \( r \) should be increased. 
    As illustrated in Fig.~\ref{fig:bar_40_combined},  as the number of rows increases, the width of each bar correspondingly expands, accompanied by a reduction in the bar's hight. In our experiments, the impact of the bar arrangements on the classification performance is investigated.   
       
    \begin{figure}[!t]
        \centerline{\includegraphics[width=0.55\columnwidth]{ 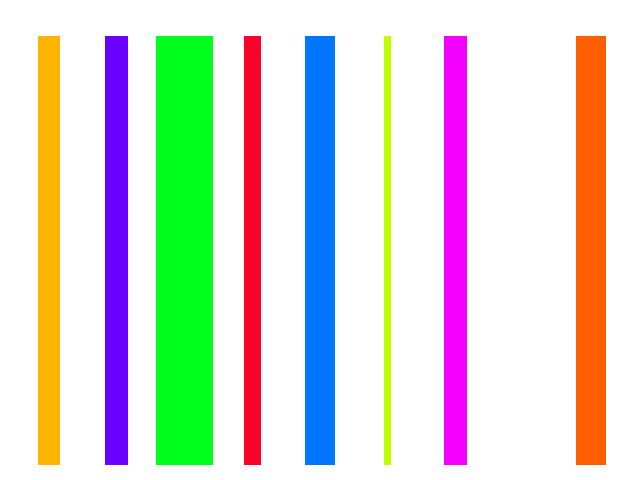}}
       % \caption{An image representation of a sample of a tabular dataset. The image consists of 9 bars of widths proportional to the feature values. The image background is white, and the 8th feature has a value of zero.} 
        \caption{Visual representation of a sample from a 9-feature tabular dataset. The bar widths are proportional to feature values. Note that the 8th feature has a value of zero, resulting in a bar with no width against the white background.}
        %Visual representation of a sample from a tabular dataset. The image depicts 9 bars, with widths proportional to the corresponding feature values. Note the image background is white, and the 8th feature has a value of zero.}
        \label{fig:galaxy9}
    \end{figure}

    \begin{figure}[h!]
        \centering
        \begin{subfigure}[b]{0.45\linewidth}
            \includegraphics[width=\linewidth]{ 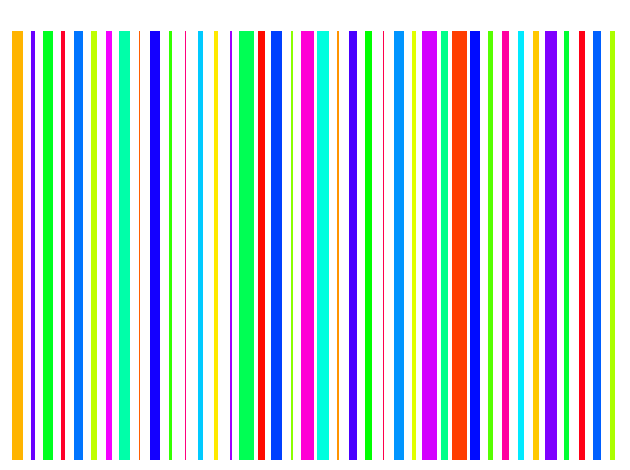}
            \caption{One row}
            \label{fig:bar_40_1}
        \end{subfigure}
        \hfill
        \begin{subfigure}[b]{0.45\linewidth}
            \includegraphics[width=\linewidth]{ 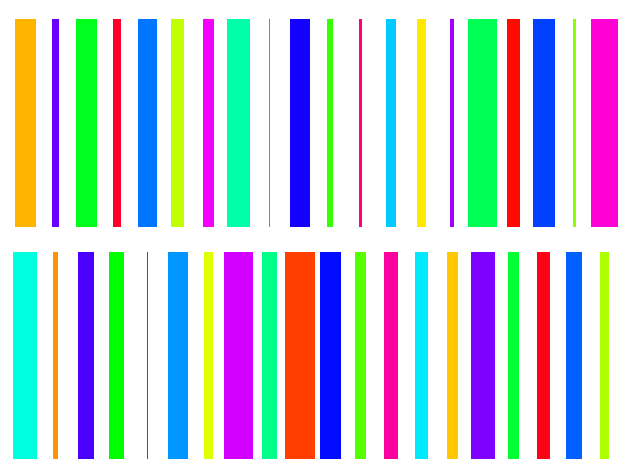}
            \caption{Two rows}
            \label{fig:bar_40_2}
        \end{subfigure}
        
        \vskip\baselineskip
        
        \begin{subfigure}[b]{0.45\linewidth}
            \includegraphics[width=\linewidth]{ 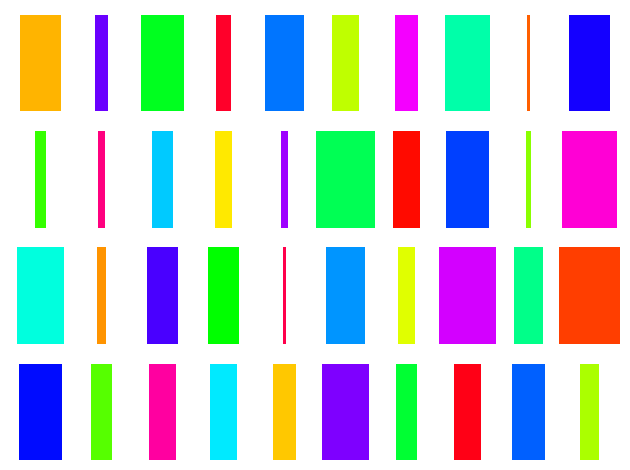}
            \caption{Four rows}
            \label{fig:bar_40_4}
        \end{subfigure}
        \hfill
        \begin{subfigure}[b]{0.45\linewidth}
            \includegraphics[width=\linewidth]{ 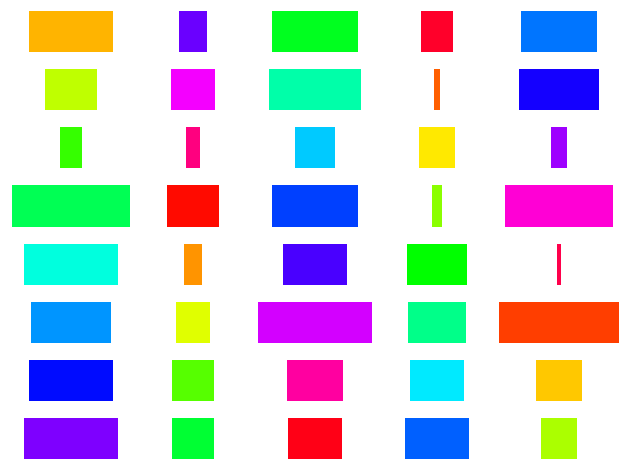}
            \caption{Eight rows}
            \label{fig:bar_40_8}
        \end{subfigure}
        
        \caption{Tab2Visual representations of a sample with 40 features arranged in different configurations.}
        \label{fig:bar_40_combined}
    \end{figure}
    
    \item \textbf{Feature Encoding:} Each feature \( x_i^{(j)} \) of the $i$-th sample is represented as a vertical bar in the image \( I_i \) having a height \( h \) and width, \( w_j \), proportional to the normalized feature value. That is, \( w_j = x_i^{(j)} \times b \).  
    A feature with a normalized value of 1 will be represented by a bar occupying the full bar support \( h \times b \) within the image.
    This representation maintains the relative feature magnitudes. Moreover, each feature is assigned a color, which enhances the visual differentiation between features and facilitates image interpretation, see Fig.~\ref{fig:galaxy9}.

    \item \textbf{Image Augmentation:} 
    This step is optional but highly recommended for smaller datasets. In such cases, a common practice in deep learning is to employ  augmentation techniques to increase the dataset size and diversity without collecting additional data samples.
    Data augmentation involves transforming the data,  thus changing the data features, while preserving its correspondence to the  label/class. Applying augmentation methods to heterogeneous tabular data presents significant challenges~\cite{Borisov_NNLS2024}. One direct advantage of representing data as images is the availability of numerous effective image augmentation techniques (e.g., mirroring, rotation, translation, cropping, intensity changing)~\cite{Xu_PR2023, Buslaev_Info2020}. Nevertheless,     
    we propose a set of image augmentation techniques specifically tailored for Tab2Visual-generated representations. Our approach incorporates elastic distortions and morphological operations, including dilation, erosion, opening, and closing, applied at varying scales. These operations are carefully selected to introduce meaningful variations to the image representations.

%our implementation of the proposed augmentation methods leverages the Albumentations library~\cite{Buslaev_Info2020}, which is a popular computer vision tool to perform a large set of fast and flexible image augmentations.
%
    More specifically, the proposed augmentation methods are implemented using the Albumentations library~\cite{Buslaev_Info2020}, 
    a fast and flexible open-source library for image augmentations.
    Initially, an image undergoes elastic distortion. 
    The extent of elastic distortion is controlled by two parameters: the scaling factor, $\alpha$, which determines the intensity of displacement, and $\sigma$, the standard deviation of the Gaussian filter applied to this field. Higher $\alpha$ values result in greater distortion, while $\sigma$ controls the nature of the distortion: lower $\sigma$ values produce small, localized ripples, while higher $\sigma$ values lead to larger, smoother wavy distortions.
    %
    %The extent of elastic distortion is regulated by two parameters: the scaling factor, $\alpha$, which controls the intensity of the displacement field, and the standard deviation, $\sigma$, of the Gaussian filter applied to the displacement field. A higher $\alpha$ value induces greater image distortion. Conversely, a lower $\sigma$ value results in small, localized, ripple-like distortions, while a higher value leads to larger, smoother, wave-like distortions.
        
Subsequently, the image undergoes random morphological operations: dilation and erosion. These operations are applied with probabilities $P_d$ and $P_e$, respectively, and with randomly sized structuring elements $SE_e$ and $SE_d$. A structuring element is a matrix that defines the neighborhood used to process each pixel.
As the size of structuring elements increases, the extent of dilation or erosion effects becomes more pronounced.
To introduce further variability, the order of dilation and erosion is randomized, leading to images that are either dilated only, eroded only, morphologically closing (dilated followed by eroded), or morphologically opening (eroded followed by dilated).
 
%Since the relevant information in the proposed image presentation primarily lies in the bar widths, these operations introduce subtle, non-uniform variations in the image's bar boundaries and widths. This augmentation strategy effectively generates synthetic samples that closely mimic the characteristics of the original data distribution and enhances the model's ability to generalize to unseen data.
%
As the critical information in the proposed image representation resides primarily in bar widths, these augmentations subtly and non-uniformly alter bar boundaries and widths. This effectively generates synthetic samples that closely resemble the original data distribution, enhancing the model's generalization ability.
Fig.~\ref{fig:galaxy10} illustrates the image representations of three samples of a dataset, each accompanied by two augmented versions.
    Note the slight to moderate nonuniform adjustments introduced by the augmentation operations along the image edges, thereby generating new samples that closely resemble the original distribution.
    
    \begin{figure}[!t]
        \centering
        \begin{minipage}{0.25\textwidth}
            \centerline{\includegraphics[width=\linewidth]{ 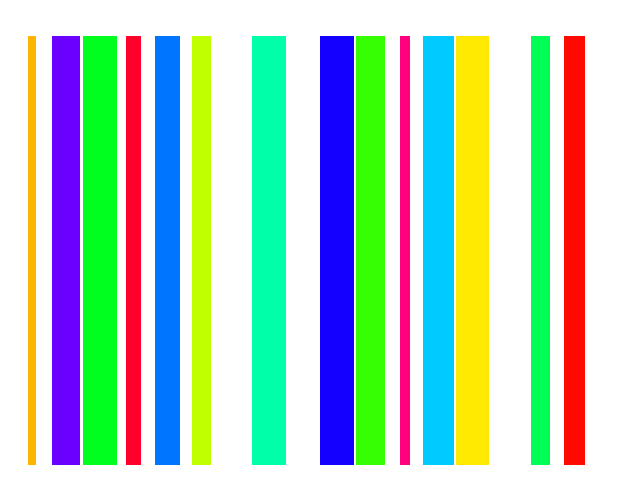}}
            \textbf{\centerline{(A)}}
        \end{minipage}
        \hfill
        \begin{minipage}{0.25\textwidth}
            \centerline{\includegraphics[width=\linewidth]{ 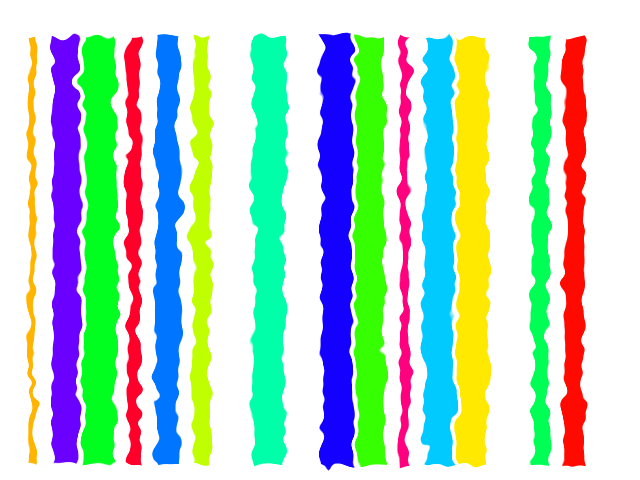}}
            \textbf{\centerline{(A')}}
        \end{minipage}
        \hfill
        \begin{minipage}{0.25\textwidth}
            \centerline{\includegraphics[width=\linewidth]{ 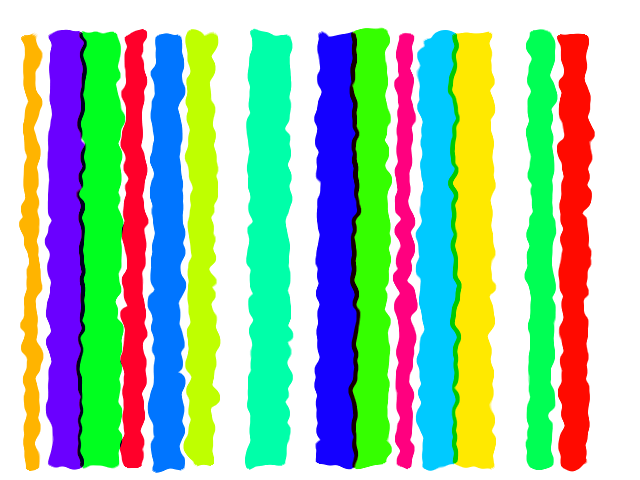}}
            \textbf{\centerline{(A'')}}
        \end{minipage}
          \hfill
        \begin{minipage}{0.25\textwidth}
            \centerline{\includegraphics[width=\linewidth]{ 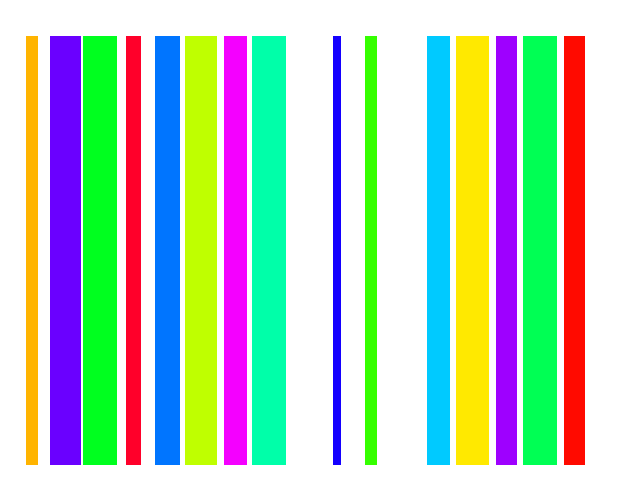}}
            \textbf{\centerline{(B)}}
        \end{minipage}
        \hfill
        \begin{minipage}{0.25\textwidth}
            \centerline{\includegraphics[width=\linewidth]{ 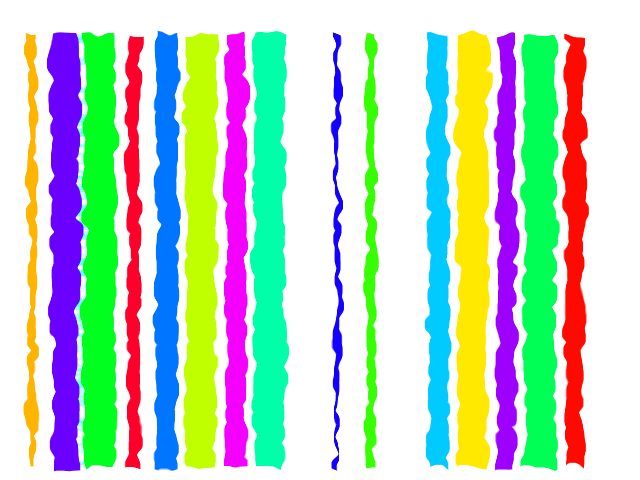}}
            \textbf{\centerline{(B')}}
        \end{minipage}
        \hfill
        \begin{minipage}{0.25\textwidth}
            \centerline{\includegraphics[width=\linewidth]{ 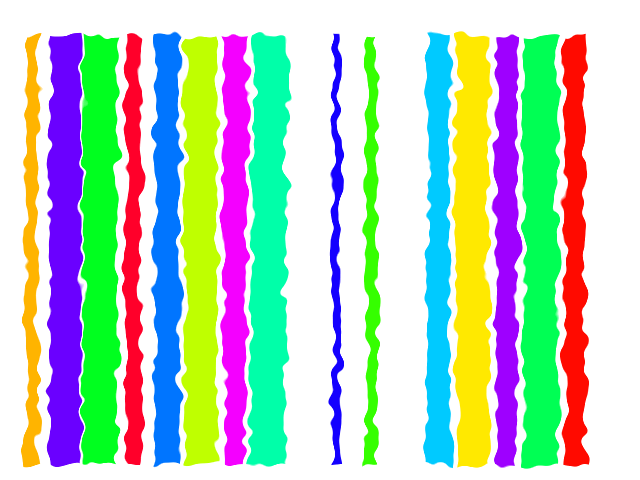}}
            \textbf{\centerline{(B'')}}
        \end{minipage}
              \hfill
        \begin{minipage}{0.25\textwidth}
            \centerline{\includegraphics[width=\linewidth]{ 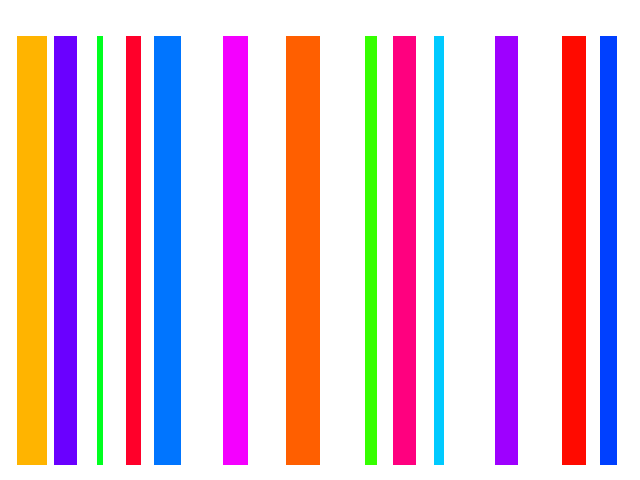}}
            \textbf{\centerline{(C)}}
        \end{minipage}
        \hfill
        \begin{minipage}{0.25\textwidth}
            \centerline{\includegraphics[width=\linewidth]{ 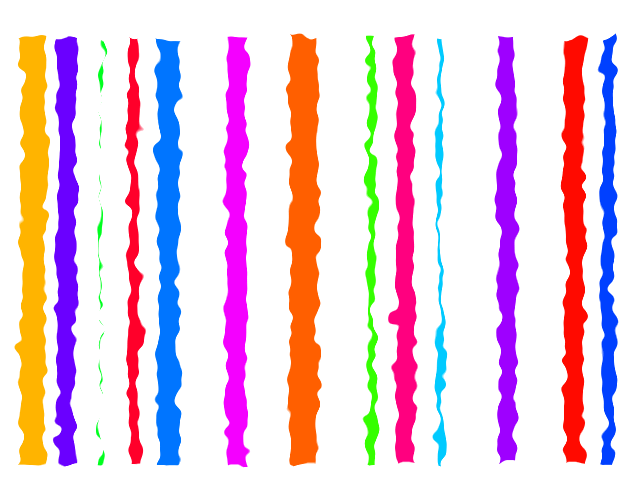}}
            \textbf{\centerline{(C')}}
        \end{minipage}
        \hfill
        \begin{minipage}{0.25\textwidth}
            \centerline{\includegraphics[width=\linewidth]{ 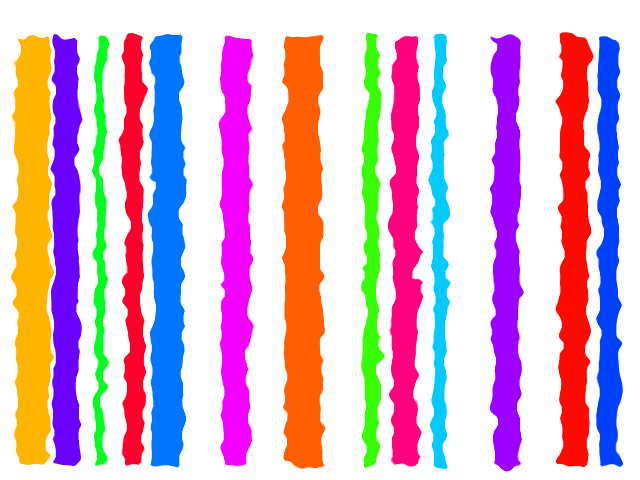}}
            \textbf{\centerline{(C'')}}
        \end{minipage}
    
         \caption{
        Example of data augmentation applied to Tab2Visual image representations of the Juice dataset. Three original samples are shown on the left, each accompanied by two augmented versions.}
        %Applying augmentation techniques to the image representation of the Juice dataset. Three samples are shown on the left with two generated image augmentations per each.}
        \label{fig:galaxy10}
    \end{figure}

    \item \textbf{Model Selection and Transfer Learning:} After image preparation and optional augmentation, the images are utilized to train a deep learning model. 
    The proposed approach offers flexibility in model selection, allowing for the use of either CNNs or ViTs depending on the specific task demands. CNNs, with their hierarchical structure, are well-suited for extracting both local and global features, while ViTs excel in capturing long-range dependencies through self-attention mechanisms. Training can be performed from scratch,learning all parameters from the dataset, or through transfer learning, where a pre-trained model is adapted to the specific task.

\end{itemize}

In summary, Algorithm~\ref{alg:alg1} outlines all the steps of the Tab2Visual approach for training a deep model $\mathcal{M}$ on a tabular dataset $\mathcal{D}$, while Algorithm~\ref{alg:alg2} details the augmentation methods of the approach. The required layout of the bars in the produced images is defined by the input parameter $r$,  whereas the desired scale of augmentation is determined by $K$.  

\begin{algorithm}
    \caption{The Tab2Visual approach for training a deep model on a tabular dataset}
     \label{alg:alg1}
    \begin{algorithmic}[1]
    \Require Tabular dataset $\mathcal{D} = \{(\textbf{x}_i, y_i)\}_{i=1}^n$ with $m$ features; Desired image dimensions $W$ and $H$; Selected number of rows $r$; Scale of augmentation $K$; Deep learning model $\mathcal{M}$.
    \Ensure Trained model $\mathcal{M}$.
    
    \State \textbf{Data Preparation and Normalization:}
    \State Encode categorical and ordinal features
    \State Apply 0-1 normalization such that $x_i^{(j)} \in [0, 1], \forall i=1,...,n, \forall j=1,...,m$

    \State \textbf{Image Preparation and Feature Encoding:}
    \State Output dataset $\mathcal{I} \leftarrow \phi$
    \State Number of columns $c \leftarrow \lceil m / r \rceil$
    \State Bar height $h \leftarrow = H / r$
    \State Maximum bar width $w \leftarrow W / c$
    
    \For{$i = 1$ to $n$}
        \State Initialize image $I_i$ of size $H \times W$ to background color
        \For{$j = 1$ to $m$}
            \State Row index $r_j \leftarrow \lceil j / c \rceil$
            \State Column index $c_j \leftarrow j - (r_j - 1) ~c$
            \State Bar width $w_j \leftarrow w \times x_i^{(j)}$
            \State Bar position:
                \State \hskip1em $x_{\text{start}} \leftarrow (c_j - 1) ~w$ % + (w - w_j)/{2}$
                \State \hskip1em $y_{\text{start}} \leftarrow (r_j - 1) ~h$
            \State Assign color $color_j$ for feature $j$
            \State Draw rectangle in $I_i$ at $(x_{\text{start}}, y_{\text{start}})$ with width $w_j$, height $h$, and color $color_j$
        \EndFor
    
        \State $\mathcal{I} \leftarrow \mathcal{I} \cup \{(I_i, y_i)\}$
        \State \textbf{Optional Image Augmentation:}
        \State {$k \leftarrow K$}
        \While{$k > 0$}
            \State $I' \leftarrow \text{AugmentImage}(I_i)$ \Comment{Apply Algorithm~\ref{alg:alg2}}
            \State $\mathcal{I} \leftarrow \mathcal{I} \cup \{(I', y_i)\}$
            \State {$k \leftarrow k-1$}
            \EndWhile
        %\State Add $(I_i, y_i)$ to the transformed dataset
    \EndFor
    
    \State \textbf{Model Training:}
    \State Train or fine-tune model $\mathcal{M}$ on the output dataset $\mathcal{I}$
\end{algorithmic}
\end{algorithm}

\begin{algorithm}
    \caption{Image Augmentation for Tab2Visual (\textbf{AugmentImage})}
     \label{alg:alg2}
    \begin{algorithmic}[1]
    \Require Image $I$; Elastic distortion parameters $\alpha$ and $\sigma$; Morphological operation probabilities $P_d$ and $P_e$;
    Structuring element sizes $S_e$ and $S_d$
    \Ensure Augmented image $I'$.
    
    \State \textbf{Elastic Distortion:}

       \State \hskip1em $I' \leftarrow \text{ApplyElasticDistortion}(I, \sigma, \alpha)$
    
    \State \textbf{Morphological Operations:}
    \State Choose two random numbers, $u$ and $v$, drawn uniformly from the interval $[0,1]$
    \State Generate structuring element $SE_d$ randomly with size $S_d$ or smaller
        \State Generate structuring element $SE_e$ randomly with size $S_e$ or smaller

    \If{$(u < P_d)$ \textbf{and} $(v < P_e)$}
        \State Choose a random number $a$ drawn uniformly from the interval $[0,1]$
         \If{$a <  0.5$}
            \State $I' \leftarrow \text{Erode}(\text{Dilate}(I', SE_d), SE_e)$ \Comment{Morphological closing}
        \Else
            \State $I' \leftarrow \text{Dilate}(\text{Erode}(I', SE_e), SE_d)$ \Comment{Morphological opening}
        \EndIf
    \ElsIf{$u < P_d$}
        \State $I' \leftarrow \text{Dilate}(I', SE_d)$
    \ElsIf{$v < P_e$}
        \State $I' \leftarrow \text{Erode}(I', SE_e)$
    \Else
        \State $I'$ remains unchanged
    \EndIf
    
    \State \textbf{Output:}
    \State Return augmented image $I'$
\end{algorithmic}
\end{algorithm}

%% Use \section commands to start a section
%%%%%%%%%%%%%%%%%%%%%%%%%%%%%%%%%%%%%%%%%%%%%%%%%%%%%%%%%%%%%%%%%%%%%%%%%%%%%%%%%%%%%%%%%5
\section{Experimental Results}
\label{sec:experiments}

This section presents a detailed experimental evaluation of Tab2Visual. We begin with an overview of the datasets used in our experiments, followed by a description of the machine learning methods employed for comparison. Next, we discuss the configuration and parameters of Tab2Visual, including the backbone deep models used for training. Finally, we report the experimental results. A detailed discussion of these results and our findings is presented in Section~\ref{sec:discussion}.

\subsection{Datasets}
Our study incorporates ten diverse datasets from the UCI Machine Learning Repository, a renowned public resource. These datasets span a range of domains, including medical, e-commerce, and engineering, providing a comprehensive evaluation ground for Tab2Visual across various application scenarios. Table~\ref{table:datasets} summarizes these datasets, which include both small-scale datasets (e.g., Heart Failure, Juice, Diabetes, Breast Cancer, Glass, US Presidential Election Results) with fewer than 1000 samples and larger-scale datasets (e.g., Satellite, Employee, Electrical Grid, Telescope) with 6000 or more samples. This diverse selection enables a thorough evaluation of Tab2Visual's generalization capabilities across different data scales.

\begin{table}[H]
  \caption{Summary of Datasets Used to Evaluate Tab2Visual.\label{table:datasets}}
  \centering
  \scriptsize
  \begin{tabularx}{\textwidth}{CCCC}
      \toprule
      \textbf{Dataset} & \textbf{Size} $(n)$ & \textbf{Attributes} $(m)$ & \textbf{Number of Classes} $(C)$\\ 
      \midrule
      Heart Failure (HRT)                    & 200      & 16   & 2 \\
      \midrule
      Juice (JU)                             & 1070     & 19   & 2 \\
      \midrule
      Diabetes (DIA)                         & 768      & 9    & 2 \\
      \midrule
      Breast Cancer (BC)                & 683      & 10   & 2 \\
      \midrule
      Glass (GL)                             & 214      & 10   & 6 \\
      \midrule
      US Presidential Election (PE) & 497 & 7  & 2 \\
      \midrule
      Satellite (SAT)                        & 6435     & 37   & 6 \\
      \midrule
      Employees (EMP)                        & 14,999   & 10   & 2 \\
      \midrule
      Electrical Grid (EG)             & 10,000   & 13   & 2 \\
      \midrule
      Telescope (TEL)                        & 19,020   & 11   & 2 \\
      \bottomrule
  \end{tabularx}
\end{table}

\subsection{Competing Machine Learning Methods}
\label{sec:ML methods}
To rigorously evaluate Tab2Visual, we compare its performance against a diverse set of classification algorithms: 
\begin{itemize}
    \item Conventional Methods: Logistic Regression, Support Vector Machines (SVM), and a shallow Multi-Layer Perceptron (MLP), which are commonly used for tabular data classification~\cite{Borisov_NNLS2024}.
    \item Tree-based Ensembles: Random Forest~\cite{ho1995random}, ExtraTrees~\cite{geurts2006extremely}, XGBoost~\cite{chen2016xgboost}, LightGBM~\cite{ke2017lightgbm}, and CatBoost~\cite{dorogush2018catboost}, renowned for their robustness and ability to capture complex feature interactions. They serve as strong baselines for tabular data classification~\cite{Borisov_NNLS2024,Grinsztajn_NIPS2022, Shwartz_IF2022}.
    \item Deep Learning Methods: state-of-the-art models specifically designed for tabular data, such as TabNet\footnote{https://github.com/dreamquark-ai/tabnet}~\cite{arik2021tabnet} and 
    TabPFN\footnote{https://github.com/PriorLabs/TabPFN. Version 1 of the software was used for this study. It is important to note that Version 2 became available during the manuscript submission process.}~\cite{hollmann2022tabpfn,Hollmann_Nature2025}, which leverage advanced techniques like attention mechanisms and probabilistic forecasting to automatically extract meaningful patterns in the data.
\end{itemize}
This comprehensive comparison provides valuable insights into the strengths and limitations of Tab2Visual relative to established and cutting-edge approaches for tabular data classification.

\subsection{Tab2Visual Configuration}
\label{tab2vis_config}
Our experiments utilize EfficientNet~\cite{tan2019efficientnet} as the initial backbone model for Tab2Visual. 
The EfficientNet series is a family of CNN architectures known for achieving state-of-the-art accuracy while maintaining high efficiency across various classification tasks~\cite{tan2019efficientnet}. This series employs a compound scaling method to uniformly scale network width, depth, and resolution, leading to a family of models (EfficientNet-B0 to B7) with varying complexity. 
The foundational EfficientNet-B0 model was developed through a synergistic approach that combined AutoML and the mobile neural architecture search framework.

EfficientNetV2~\cite{tan2021efficientnetv2} builds on its predecessor by integrating training-aware neural architecture search with scaling strategies, resulting in enhanced efficiency and accuracy. Notable architectural modifications include the removal of depthwise convolutions and squeeze-and-excitation blocks in the early layers, along with the use of smaller kernel sizes and contraction ratios in mobile blocks. Additionally, an improved progressive learning technique accelerates training while boosting accuracy.
Empirical studies~\cite{tan2021efficientnetv2, deng2022deepfake, banerjee2023ceimven} have highlighted EfficientNetV2's strong performance in transfer learning tasks, achieving high accuracy with fewer parameters than competing models. These characteristics make it particularly suitable for transfer learning, especially in scenarios with limited data.
To minimize overfitting and computational costs, we opt for the smallest variant, EfficientNetV2-B0, 
which is pre-trained on ImageNet-1K~\cite{rw2019timm}, a labeled dataset comprising 1.2 million images spanning 1,000 categories.

In Tab2Visual experiments, we set the various parameters as follows. We select $r=1$ for all datasets, arranging the bars in one row in the produced images.  We set the image height $H$ and width $W$ to 224 pixels each. This choice aligns with the input size requirement of EfficientNetV2, which is designed to accept images of $224 \times 224$ pixels. 
For the smaller datasets, namely (heart, glass, diabetes, breast cancer, USA election and juice), we apply the proposed set of augmentation methods (see Algorithm~\ref{alg:alg2}) to the training partition of each dataset.  We experiment with different control parameter settings. 
For the elastic distortion part, we use $\alpha \in [40, 60]$, and $\sigma \in [3, 5]$. In the morphological operation part,
we employ $P_d \in [0.6, 0.8]$, and $P_e \in [0.6, 0.8]$. Both the structuring elements $SE_e$ and $SE_d$ are randomly set to size $(2, 5)$ or smaller. 

% %%now talk about training
To mitigate overfitting, particularly when dealing with small datasets, we implement a multi-faceted strategy during backbone model training. 
Firstly, we apply the proposed image augmentation methods to small datasets to expand the effective dataset size. 
Secondly, we utilize EfficientNetV2-B0, the smallest and least complex model in the EfficientNetV2 family, which inherently minimizes the risk of overfitting owing to its smaller number of parameters.
Thirdly, we employ transfer learning, in which, the pre-trained weights of the deep model are frozen, except for the weights of the final classification layer. This allows the model to adapt to the specific classification task while preserving learned feature representations. 
Finally, we incorporate a combination of regularization techniques:
\begin{itemize}
    \item Batch Normalization: This technique stabilizes training and reduces internal covariate shift, allowing the model to learn more effectively across different mini-batches.
    \item Strong Weight Decay (L2 Regularization): By penalizing large weights, this technique promotes simpler models that are less prone to overfitting.
    \item Dropout: Randomly deactivating neurons during training reduces reliance on individual neurons and encourages more robust feature representations.
\end{itemize}
This multi-faceted approach effectively mitigates overfitting, enabling our model to generalize well to unseen data.

\subsection{Experimental Setup and Results}
\label{results_config}
All methods mentioned in Section~\ref{sec:ML methods}, in addition to Tab2Visual, are implemented in Python and evaluated using 5-fold cross-validation. Two metrics are used to assess a method's performance: the macro average F1-score as well as the Area Under the Receiver Operating Characteristic Curve (AUC). The former metric measures the harmonic mean of precision and recall of a method, providing a balanced assessment, while the latter offers a measure of a method's average performance across different 
trade-offs between true positive and false positive rates at various thresholds. 

On using augmentation, data is first divided into non-overlapping training and testing partitions. Augmentation is then applied exclusively to the training partition to prevent data leakage. This ensures that augmented data from the training partition do not inadvertently influence the evaluation on the held-out test partition, maintaining the integrity and generalizability of the evaluation process.
Different augmentation scales are investigated (see Table~\ref{table:scales}) to evaluate the influence of augmentation on model performance.

\begin{table} [H]
    \centering
     \caption{Augmentation Scales.\label{table:scales}}
     \scriptsize
    \begin{tabular}{|c|c|l|}
    \hline
       $\mathbf{K}$  & \textbf{Augmented Data} & \textbf{Description}\\
       \hline
       0  &  \( \mathcal{A}_0 \) & No augmentations applied\\
       1  & \( \mathcal{A}_1 \) & Each image augmented once \\
       2  & \( \mathcal{A}_2 \) & Each image augmented twice\\
       3  & \( \mathcal{A}_3 \) & Each image augmented three times\\
       4  & \( \mathcal{A}_4 \) & Each image augmented four times\\
       \hline
    \end{tabular}
\end{table}

To ensure fair comparison, we perform hyperparameter optimization for all methods using the Optuna~\cite{akiba2019optuna} framework with 100 iterations. Each hyperparameter configuration is cross-validated with the 5 folds. 
Table~\ref{table:2} summarizes the hyperparameters and their search spaces for each method, including the best-performing hyperparameter settings for the US Presidential Election Results dataset. For details on hyperparameter abbreviations, please refer to the scikit-learn documentation~\cite{scikit-learn}. 
%All experiments are conducted on a workstation equipped with an Intel Core i9 CPU running at 3 MHz, featuring 18 cores, 256 GB of RAM, and an NVidia Quadro RTX 5000 GPU with 12 GB of dedicated memory. 
All experiments are performed on a workstation with an Intel Core i9 CPU (3 GHz, 18 cores), 256 GB RAM, and an NVIDIA Quadro RTX 5000 GPU (12 GB VRAM).

Table~\ref{table:Tab2Visual_ablation} summarizes average F1-score and AUC results for all methods across all datasets. 
Note image augmentation is applied exclusively to datasets with fewer than 1000 samples.
Therefore, Tab2Visual results are not reported for augmented data ($\mathcal{A}_k$, for $k>0$) for larger datasets. In contrast, there are 5 variants of the Tab2Visual model on smaller datasets (one for each augmentation level). Additionally, due to the method's limitations as noted in the original paper~\cite{hollmann2022tabpfn}, TabPFN results are not reported for datasets larger than 1000 samples.

\begin{table}[p]  %H
    \centering                 % centers the table horizontally
    \caption{Optimizing hyperparameters for all classification models.
    %Hyperparameter tuning for all classification methods.
    \label{table:2}}
    \tiny
    \begin{adjustwidth}{0cm}{0cm} % Replace \extralength with a fixed value
        \begin{tabularx}{\textwidth}{CCCC} % Replace \fulllength with \textwidth
            \toprule
            \textbf{Model} & \textbf{Hyperparameter} & \textbf{Search Range} & \textbf{Best Configuration} \\
            \midrule
                \multirow[m]{2}{*}{Logistic Regression}	& penalty & [l1, l2] & l2 \\
                                                        & C & 0.01 to 1 & 0.833 \\
                                    \midrule
                \multirow[m]{3}{*}{SVM}         & C & 0.001 to 100 & 38.81 \\
                                                & kernel & [linear, poly, rbf] & rbf \\
                                                & gamma & 0.001 to 100 & 0.01 \\
                                    \midrule
                \multirow[m]{5}{*}{Random Forest}    & n\_estimators & 1 to 500 & 100 \\
                                                        & max\_depth & 1 to 40 & 9 \\
                                                        & min\_samples\_split & 2 to 14 & 7 \\
                                                        & min\_samples\_leaf & 1 to 14 & 4 \\
                                                        & max\_features & [auto, sqrt, log2] & auto\\
                                    \midrule
            \multirow[m]{5}{*}{Extra Trees}             & n\_estimators & 1 to 500 & 150 \\
                                                        & max\_depth & 1 to 40 & 5 \\
                                                        & min\_samples\_split & 2 to 14 & 10 \\
                                                        & min\_samples\_leaf & 1 to 14 & 5 \\
                                                        & max\_features & [auto, sqrt, log2] & auto\\
                                    \midrule
            \multirow[m]{8}{*}{XGBoost}                 & n\_estimators & 1 to 500 & 300 \\
                                                        & max\_depth & 1 to 40 & 10 \\
                                                        & gamma & 0 to 1 & 0.011 \\
                                                        & learning\_rate & 0.001 to 1 & 0.05 \\
                                                        & reg\_alpha & 0 to 2 & 0.814 \\
                                                        & reg\_lambda & 0 to 2 & 1.478 \\
                                                        & subsample & 0.5 to 1 & 0.614 \\
                                                        & colsample\_bytree & 0.5 to 1 & 0.7 \\
                                    \midrule      
                \multirow[m]{8}{*}{LightGBM}        & n\_estimators & 1 to 500 & 260 \\
                                                        & max\_depth & 1 to 20 & 7 \\
                                                        & num\_leaves & 2 to 256 & 70 \\
                                                        & learning\_rate & 0.01 to 1 & 0.1 \\
                                                        & reg\_alpha & 0 to 2 & 0.11 \\
                                                        & reg\_lambda & 0 to 2 & 1.03 \\
                                                        & subsample & 0.5 to 1 & 0.92 \\
                                                        & colsample\_bytree & 0.5 to 1 & 0.81 \\
                                    \midrule  
                \multirow[m]{4}{*}{CatBoost}        & iterations & 50 to 300 & 100 \\
                                                        & learning\_rate & 0.01 to 0.3 & 0.12 \\
                                                        & depth & 2 to 12 & 6 \\
                                                        & l2\_leaf\_reg & 1 to 10 & 3.46 \\
                                    \midrule  
                \multirow[m]{6}{*}{MLP}     & n\_hidden\_layers & [1, 2, 3] & 2 \\
                                                        & n\_neurons\_hidden\_layer & [32, 64, 128, 256] & [128, 64] \\
                                                        & learning\_rate & 0.0001 to 0.1 & 0.001 \\
                                                        & batch\_size & [16, 32, 64, 128] & 16 \\
                                                        & weight\_decay & 0.00001 to 0.01 & 0.002 \\
                                                        & drop\_prob & 0.1 to 0.7 & 0.4 \\
                                    \midrule  
                \multirow[m]{10}{*}{TabNet}        & batch\_size & [8, 16, 32, 64] & 8 \\
                                                        & mask\_type & [entmax, sparsemax] & entmax \\
                                                        & n\_d & 8 to 64 (step 4) & 8 \\
                                                        & n\_a & 8 to 64 (step 4) & 8 \\
                                                        & n\_steps & 1 to 8 (step 1) & 2 \\
                                                        & gamma & 1.0 to 1.4 (step 0.2) & 1.291 \\
                                                        & n\_shared & 1 to 3 & 1 \\
                                                        & lambda\_sparse & 0.0001 to 1 & 0.003 \\
                                                        & patienceScheduler & 3 to 10 & 6 \\
                                                        & learning\_rate & 0.001 to 1 & 0.023 \\ 
                                    \midrule   
                \multirow[m]{1}{*}{TabPFN}           & n\_ensemble\_configurations & [1, 32] & 4 \\  
                                    \midrule  
                \multirow[m]{6}{*}{Tab2Visual}     & n\_hidden\_layers & [1, 2, 3] & 1 \\
                                                        & n\_neurons\_hidden\_layer & [32, 64, 128, 256] & 128 \\
                                                        & learning\_rate & 0.000001 to 1 & 0.0005 \\
                                                        & batch\_size & [16, 32, 64, 128] & 32 \\
                                                        & weight\_decay & 0.00001 to 0.5 & 0.005 \\
                                                        & drop\_prob & 0.1 to 0.7 & 0.5 \\
                \bottomrule
        \end{tabularx}
    \end{adjustwidth}
\end{table}

%performance table
\begin{table*}[ht]
    \caption{Performance comparison of Tab2Visual against other machine learning methods on different datasets.}
    \tiny 
    \label{table:Tab2Visual_ablation}
    \begin{adjustwidth}{0cm}{0cm}  
        \centering
        { \begin{tabularx}{\fulllength}{>{\tiny}l *{11}{>{\tiny}Y}}
          %\begin{tabularx}{\textwidth}{>{\scriptsize}l *{11}{Y}}

            \toprule
            \textbf{} & 
            \textbf{} & 
             \textbf{HRT}
            & \textbf{JU}
            & \textbf{DIA}
            & \textbf{BC}
            & \textbf{GL}
            & \textbf{PE}
            & \textbf{SAT}
            & \textbf{EG}
            & \textbf{EMP}
            & \textbf{TEL} \\
            \midrule
            \textbf{Logistic Regression} & 
            F1  AUC & 
            0.5131  0.7216 & 
            0.7555  0.8907 & 
            0.6174  0.8017 & 
            0.9545  0.9914 & 
            0.4010  0.8011 & 
            0.9261  0.9811 & 
            0.7336  0.9600 & 
            0.8599  0.8922 & 
            0.7560  0.8382 & 
            0.6711  0.8328 \\
            \midrule
            \textbf{SVM} & 
            F1  AUC & 
            0.4762  0.6512 & 
            0.6814  0.8147 & 
            0.5861  0.6836 & 
            0.9432  0.9872 & 
            0.3396  0.7803 & 
            0.9173  0.9514 & 
            0.7137  0.9432 & 
            0.8140  0.8805 & 
            0.7369  0.8157 & 
            0.7260  0.7887 \\
            \midrule
            \textbf{Random Forest} & 
            F1  AUC & 
            0.4444  0.7327 & 
            0.7105  0.8666 & 
            0.5722  0.7938 & 
            0.9423  0.9937 & 
            0.6470  0.9209 & 
            0.9238  0.9763 & 
            0.8863  0.9905 & 
            0.9361  0.9774 & 
            \textbf{0.9748}  0.9913 & 
            0.8091  0.9307 \\
            \midrule
            \textbf{Extra Trees} & 
            F1  AUC & 
            0.4328  0.7287 & 
            0.6930  0.8426 & 
            0.5592  0.7782 & 
            0.9465  0.9947 & 
            \textbf{0.6874}  \textbf{0.9329} & 
            0.9094  0.9628 & 
            0.8914  0.9914 & 
            0.9391  0.9813 & 
            0.9665  0.9899 & 
            0.7976  0.9289 \\
            \midrule
            \textbf{XGBoost} & 
            F1  AUC & 
            0.4796  0.7333 & 
            0.7160  0.8648 & 
            0.5765  0.7639 & 
            0.9318  0.9916 & 
            0.6515  0.8901 & 
            0.9204  0.9698 & 
            0.8875  0.9911 & 
            0.9544  0.9876 & 
            0.9694  0.9926 & 
            \textbf{0.8194}  \textbf{0.9353} \\
            \midrule
            \textbf{LightGBM} & 
            F1  AUC & 
            0.4444  0.6982 & 
            0.7228  0.8681 & 
            0.5553  0.7665 & 
            0.9434  0.9948 & 
            0.5740  0.9086 & 
            0.9113  0.9690 & 
            0.8904  0.9913 & 
            0.9497  0.9857 & 
            0.9697  \textbf{0.9932} & 
            0.8150  0.9320 \\
            \midrule
            \textbf{CatBoost} & 
            F1  AUC & 
            0.4991  0.7754 & 
            0.7322  0.8839 & 
            0.5722  0.7918 & 
            0.9435  0.9943 & 
            0.6577  0.9290 & 
            0.9212  0.9790 & 
            \textbf{0.8942}  \textbf{0.9925} & 
            0.9599  0.9906 & 
            0.9628  0.9919 & 
            0.8131  \textbf{0.9353} \\
            \midrule
            \textbf{MLP} & 
            F1  AUC & 
            0.5049  0.6544 & 
            0.7707  0.8919 & 
            0.6368  0.8158 & 
            0.9615  0.9936 & 
            0.5178  0.7947 & 
            0.9381  0.9767 & 
            0.8612  0.9828 & 
            0.9336  0.9714 & 
            0.9079  0.9750 & 
            0.7858  0.9148 \\
            \midrule
            \textbf{TabNet} & 
            F1  AUC & 
            0.4976  0.6778 & 
            \textbf{0.7837}  0.8887 & 
            0.6205  0.7995 & 
            0.9609  0.9949 & 
            0.5156  0.7834 & 
            \textbf{0.9409}  0.9808 & 
            0.8837  0.9863 & 
            \textbf{0.9724}  \textbf{0.9950} & 
            0.9447  0.9844 & 
            0.8066  0.9302 \\
            \midrule
            \textbf{TabPFN} & 
            F1  AUC & 
            0.4287  0.7193 & 
            0.7658  0.8944 & 
            0.6321  \textbf{0.8216} & 
            0.9614  0.9925 & 
            0.5349  0.8264 & 
            0.9333  0.9829 & 
            --- & 
            --- & 
            --- & 
            --- \\
            \midrule
            \textbf{Tab2Visual \( \mathcal{A}_0 \)} & 
            F1  AUC & 
            0.5246  0.7346 & 
            0.7510  0.8672 & 
            0.6262  0.7790 & 
            0.9561  0.9884 & 
            0.6014  0.8604 & 
            0.9331  0.9655 & 
            0.8761  0.9826 & 
            0.9353  0.9742 & 
            0.9513  0.9830 & 
            0.7823  0.9100 \\
            \midrule
            \textbf{Tab2Visual \( \mathcal{A}_1 \)} & 
            F1  AUC & 
            0.5014  0.7908 & 
            0.7428  0.8288 & 
            0.6227  0.7693 & 
            0.9591  0.9872 & 
            0.6182  0.8817 & 
            0.9204  0.9704 & 
            --- & 
            --- & 
            --- & 
            --- \\
            \midrule
            \textbf{Tab2Visual \( \mathcal{A}_2 \)} & 
            F1  AUC & 
            0.5179  0.7374 & 
            0.7402  0.8710 & 
            0.6211  0.7809 & 
            0.9501  0.9880 & 
            0.6119  0.8822 & 
            0.9148  0.9759 & 
            --- & 
            --- & 
            --- & 
            --- \\
            \midrule
            \textbf{Tab2Visual \( \mathcal{A}_3 \)} & 
            F1  AUC & 
            0.4962  0.7601 & 
            0.7488  0.8596 & 
            0.6291  0.7891 & 
            \textbf{0.9647}  \textbf{0.9954} & 
            0.6469  0.8892 & 
            0.9260  0.9512 & 
            --- & 
            --- & 
            --- & 
            --- \\
            \midrule
            \textbf{Tab2Visual \( \mathcal{A}_4 \)} & 
            F1  AUC & 
            \textbf{0.5390}  \textbf{0.7755} & %was 0.7725
            0.7675  \textbf{0.8957} & 
            \textbf{0.6440}  0.7873 & 
            0.9562  0.9906 & 
            0.6396  0.8927 & 
            0.9270  \textbf{0.9841} & 
            --- & 
            --- & 
            --- & 
            --- \\
            \bottomrule
        \end{tabularx}}
    \end{adjustwidth}
\end{table*}

%%%%%%%%%%%%%%%%%%%%%%%%%%%%%%%%%%%%%%%%%%%%%%%%%%%%%%%%%%%%%%%%%%%%%%%%%%%%%%%%%%%%%%%%%%%%%%%%%%%%%%%%%%%%%%%%%%%%%%%%%%%%
\section{Discussion} 
\label{sec:discussion}
We here present a detailed comparative analysis of the accuracy and speed of Tab2Visual against other methods. We evaluate the performance of Tab2Visual across various datasets, examining its strengths and limitations in different scenarios. Furthermore, we perform an in-depth analysis of each key component of the Tab2Visual approach, including the impact of image augmentation, the benefit from transfer learning, the choice of backbone model, and the influence of different image feature arrangements.

\subsection{Performance Comparison on Different Datasets}
Our experimental evaluation reveals varying performance across datasets, with no universally superior method.
On smaller datasets like Heart, Juice, Diabetes, Breast Cancer, Glass, and US Election, Tab2Visual model variants (particularly from \( \mathcal{A}_4 \) and \( \mathcal{A}_3 \)) demonstrate strong performance, achieving high F1-scores and AUC values. More specifically, Tab2Visual-$\mathcal{A}_4$ achieves the highest F1-score in the Heart and Diabetes datasets, indicating good balance between precision and recall on both datasets. It also achieves the highest AUC in the Heart, Juice and USA Election datasets. Tab2Visual-$\mathcal{A}_3$ stands out with superior F1-score and AUC metrics in the Breast Cancer dataset.

Advanced deep learning models also demonstrate strong performance. TabNet excels in the Juice and USA Election datasets, achieving the highest F1-scores in both. 
TabNet exhibits the highest F1-score and AUC in the Electrical Grid dataset, while TabPFN achieves the highest AUC in the Diabetes dataset.
Tree-based ensembles, such as CatBoost, ExtraTrees, XGBoost, and LightGBM, consistently exhibit robust performance. ExtraTrees achieves the highest F1-score and AUC in the Glass dataset. In the Satellite dataset, CatBoost demonstrates superior performance, while XGBoost excels in the Telescope dataset. While traditional models like Logistic Regression, MLP, and SVM show commendable results, they are often outperformed by the other methods across the various datasets.

\begin{figure}
    \centerline{\includegraphics[width=\linewidth]{ 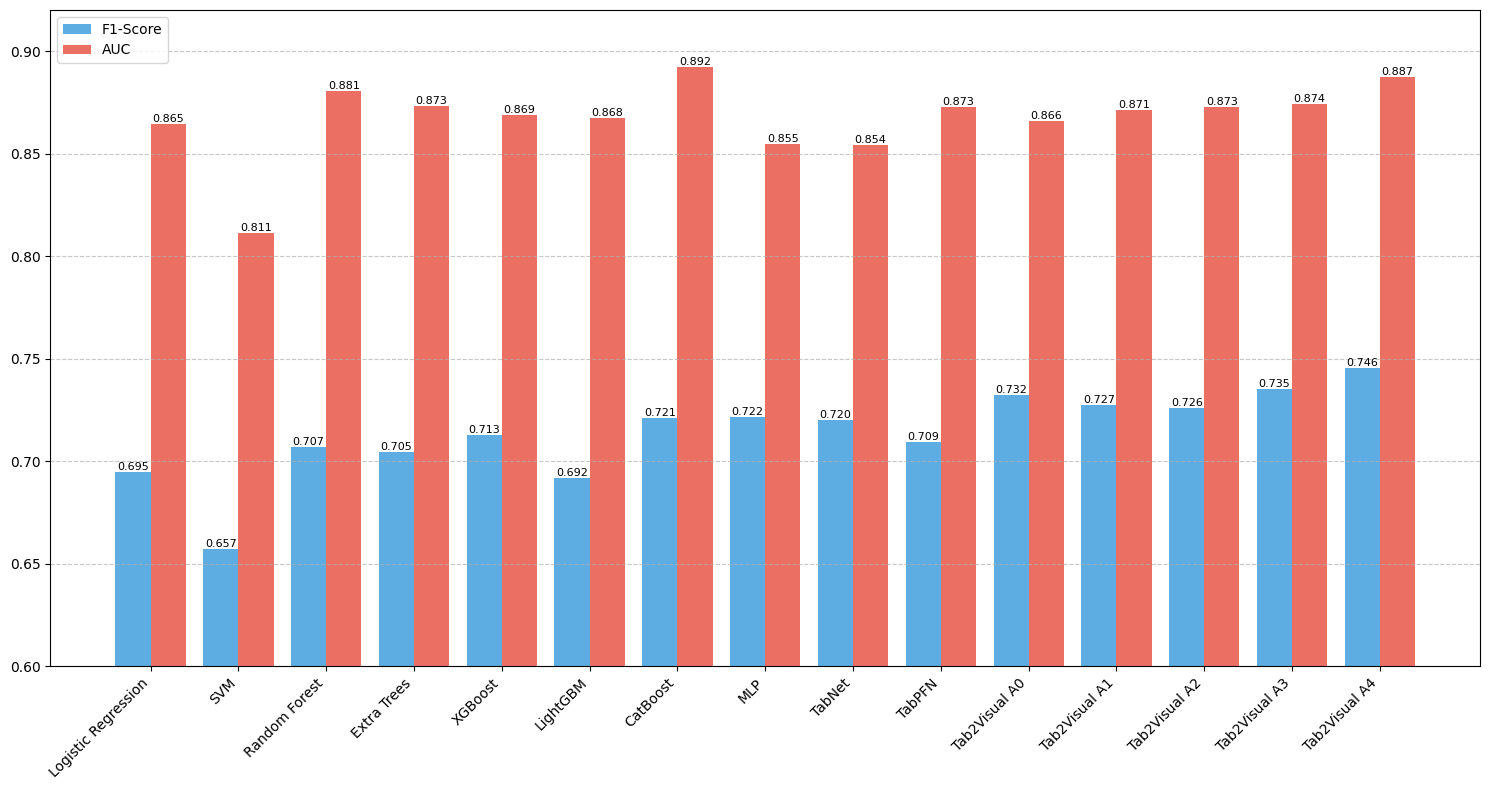}}
    \caption{Average F1-score and AUC of different classification algorithms on smaller datasets ($\leq$ 1000 samples) as evaluated in our experiments. Higher values indicate better performance.}
    %The performance in terms of average F1-score and average AUC of various classifiers over all small datasets ($<$ 1000 samples) in our experiments (the higher, the better).}
    \label{fig:bar_small}
\end{figure}

\begin{figure}
    \centerline{\includegraphics[width=\linewidth]{ 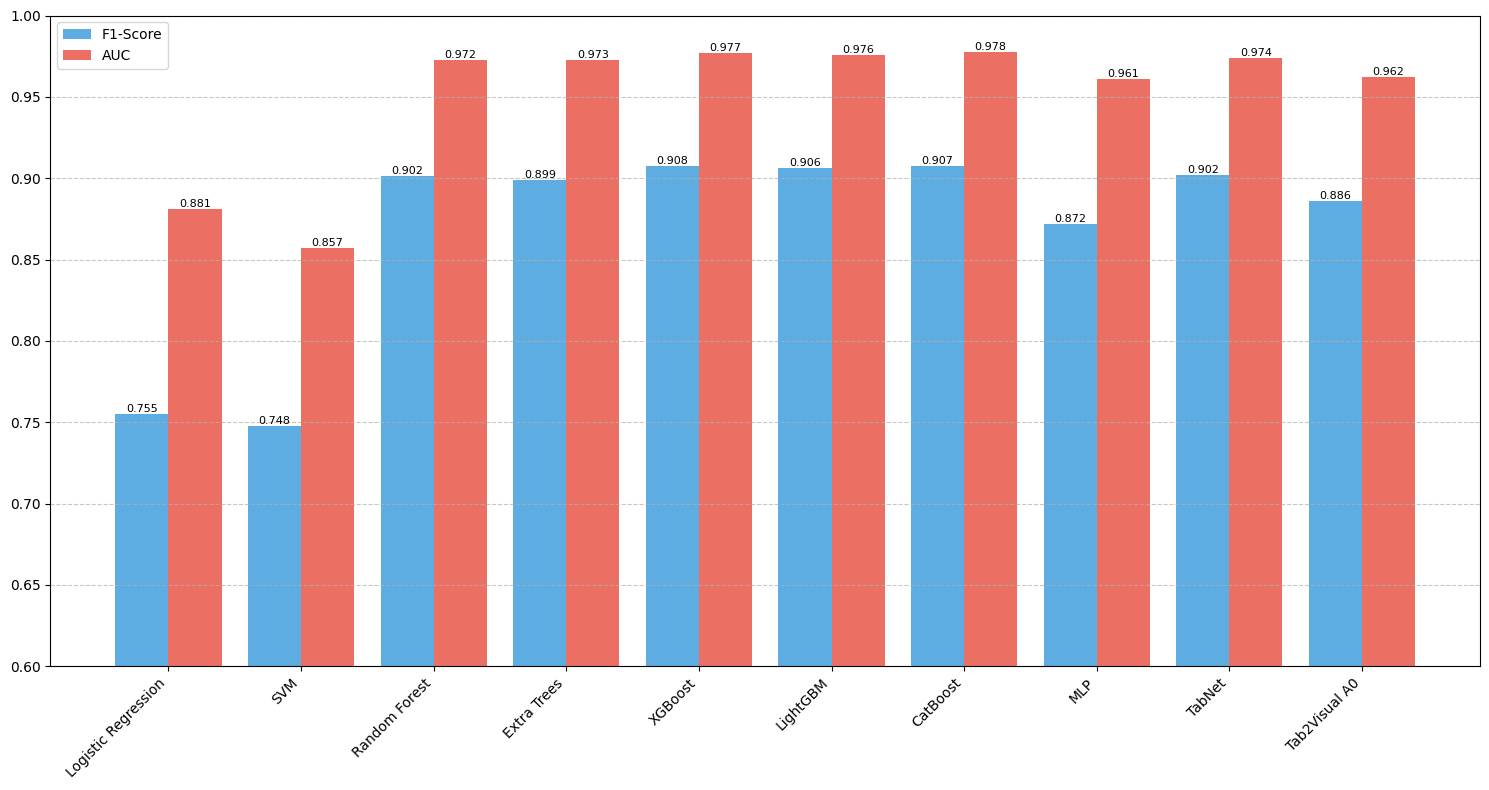}}
    \caption{Average F1-score and AUC of different classification algorithms on larger datasets ($\ge$ 6000 samples) as evaluated in our experiments. Higher values indicate better performance.}
    %The performance in terms of average F1-score and average AUC of various classifiers over all large datasets ($>$ 6000 samples) in our experiments (the higher, the better).}
    \label{fig:bar_large}
\end{figure}

\begin{figure}
    \centerline{\includegraphics[width=\linewidth]{ 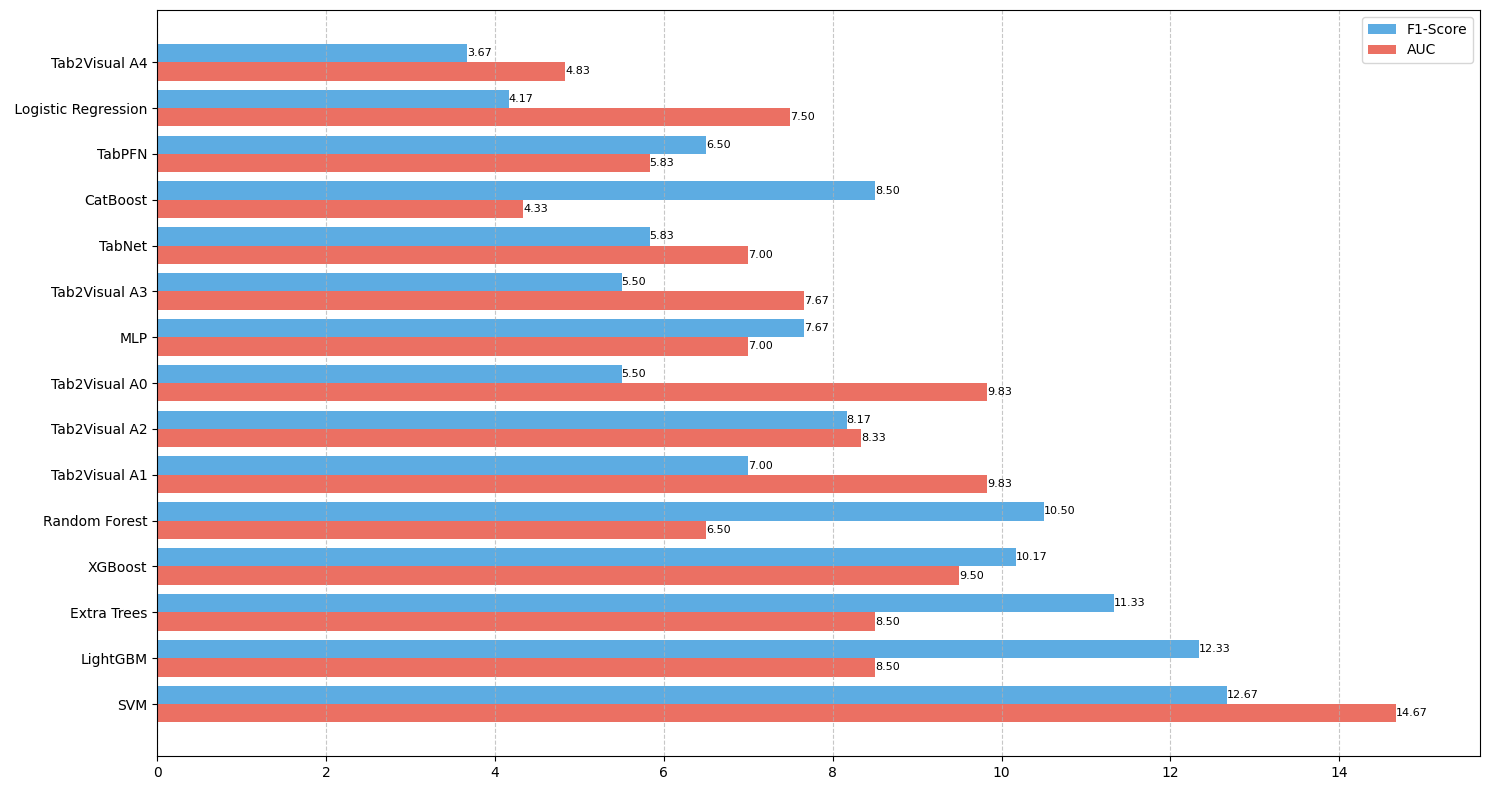}}
    \caption{Average rank of classification algorithms based on F1-score and AUC across smaller datasets ($\leq$ 1000 samples) as evaluated in our experiments. Lower ranks indicate better overall performance.}
    %The average rank based on F1-score and AUC of various classifiers over small datasets ($<$ 1000 samples) in our experiments (the smaller, the better).}
    \label{fig:ranks_small}
\end{figure}

\begin{figure}
    \centerline{\includegraphics[width=\linewidth]{ 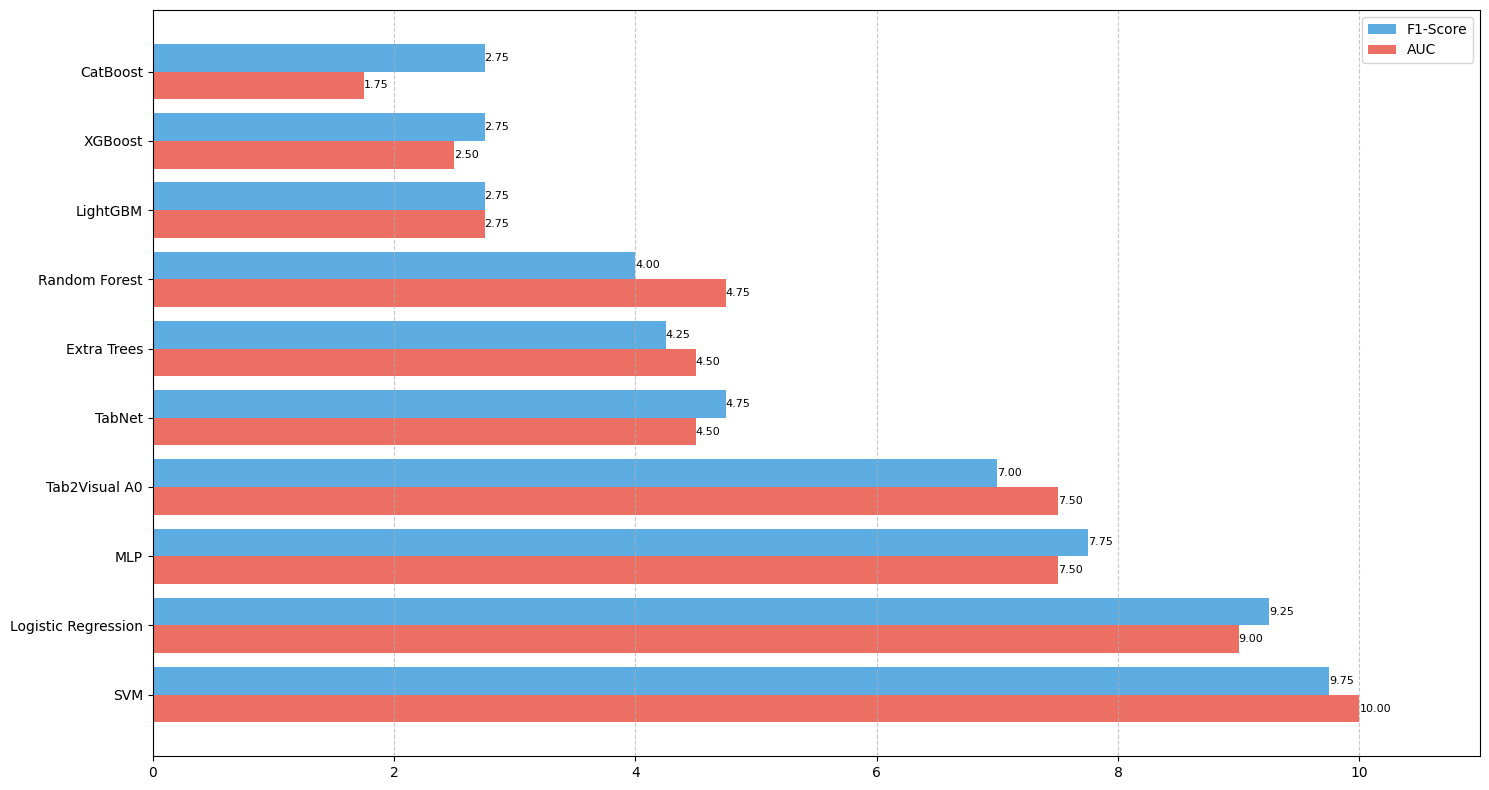}}
    \caption{Average rank of classification algorithms based on F1-score and AUC across larger datasets ($\ge$ 6000 samples) as evaluated in our experiments. Lower ranks indicate better overall performance.}
    %The average rank based on F1-score and AUC of various classifiers over large datasets ($>$ 6000 samples) in our experiments (the smaller, the better). }
    \label{fig:ranks_large}
\end{figure}

While no single method consistently outperforms all others across all datasets, distinct performance trends emerge when comparing results on smaller ($\leq$ 1000 samples) and larger ($\ge$ 6000 samples) datasets. Figures~\ref{fig:bar_small} and~\ref{fig:bar_large} illustrate the average F1-score and AUC performance of each method on these two dataset categories, respectively. Furthermore, Figures~\ref{fig:ranks_small} and~\ref{fig:ranks_large} depict the average ranks of the competing methods on smaller and larger datasets based on F1-score and AUC. These figures clearly demonstrate the varying performance trends of different methods across different dataset sizes.

Our analysis reveals a clear advantage for Tab2Visual on smaller datasets. Tab2Visual models, particularly $\mathcal{A}_4$ and $\mathcal{A}_3$, consistently achieve high average F1-scores and AUC values on these datasets. In fact, Tab2Visual-$\mathcal{A}_4$ achieves the highest average F1-score of 74.6\% in Figure~\ref{fig:bar_small}, indicating strong performance in balancing precision and recall. 
It surpasses second-place methods (CatBoost, MLP, and TabPFN) by at least 2\%, with a notable 4\% improvement over TabPFN.
CatBoost and  Tab2Visual-$\mathcal{A}_4$ achieve the highest average AUC on small datasets, with CatBoost exhibiting a slight performance edge (89.2\% vs. 88.7\%). 
Significantly,  Tab2Visual-$\mathcal{A}_4$ emerges as the best average rank among all methods, as demonstrated in Figure~\ref{fig:ranks_small}.
Conversely, on larger datasets, tree-based ensembles show very close performance, achieving top positions in terms of average AUC (about 97\%), F1-score (about 90\%), and rank (see Figures~\ref{fig:bar_large} and~\ref{fig:ranks_large}). TabNet then Tab2Visual follow closely behind in terms of performance on larger datasets. SVM shows the least performances on both smaller and larger datasets.

These results demonstrate that Tab2Visual exhibits a distinct advantage in handling limited data, consistently outperforming other methods on smaller datasets. This performance gain can be attributed to several factors. Tab2Visual's image augmentation techniques effectively enhance data diversity and size, particularly beneficial for small datasets. Moreover, the EfficientNet architecture within Tab2Visual, combined with transfer learning, leverages knowledge learned from larger, pre-trained datasets, compensating for the limited availability of training data.
Conversely, on larger datasets, the benefits of data augmentation may be less pronounced due to the inherent diversity within the data itself. This allows tree-based ensembles, which excel at capturing complex relationships within larger datasets, to demonstrate superior performance.

%%%%%%%%%%%%%%%%%%%%%%%%%%%%%%%%%%%%%%%%%%%%%%%%%%%%%%%%%%%%%%%%%%%%%%%%%%%%%%%%%%%%%%%%%%%%%%%%%%%%%%%%%%%%%%%%%%%
\subsection{Augmentation Effect}
\label{sec:augmentation}
\begin{figure}
    \centerline{\includegraphics[width=\linewidth]{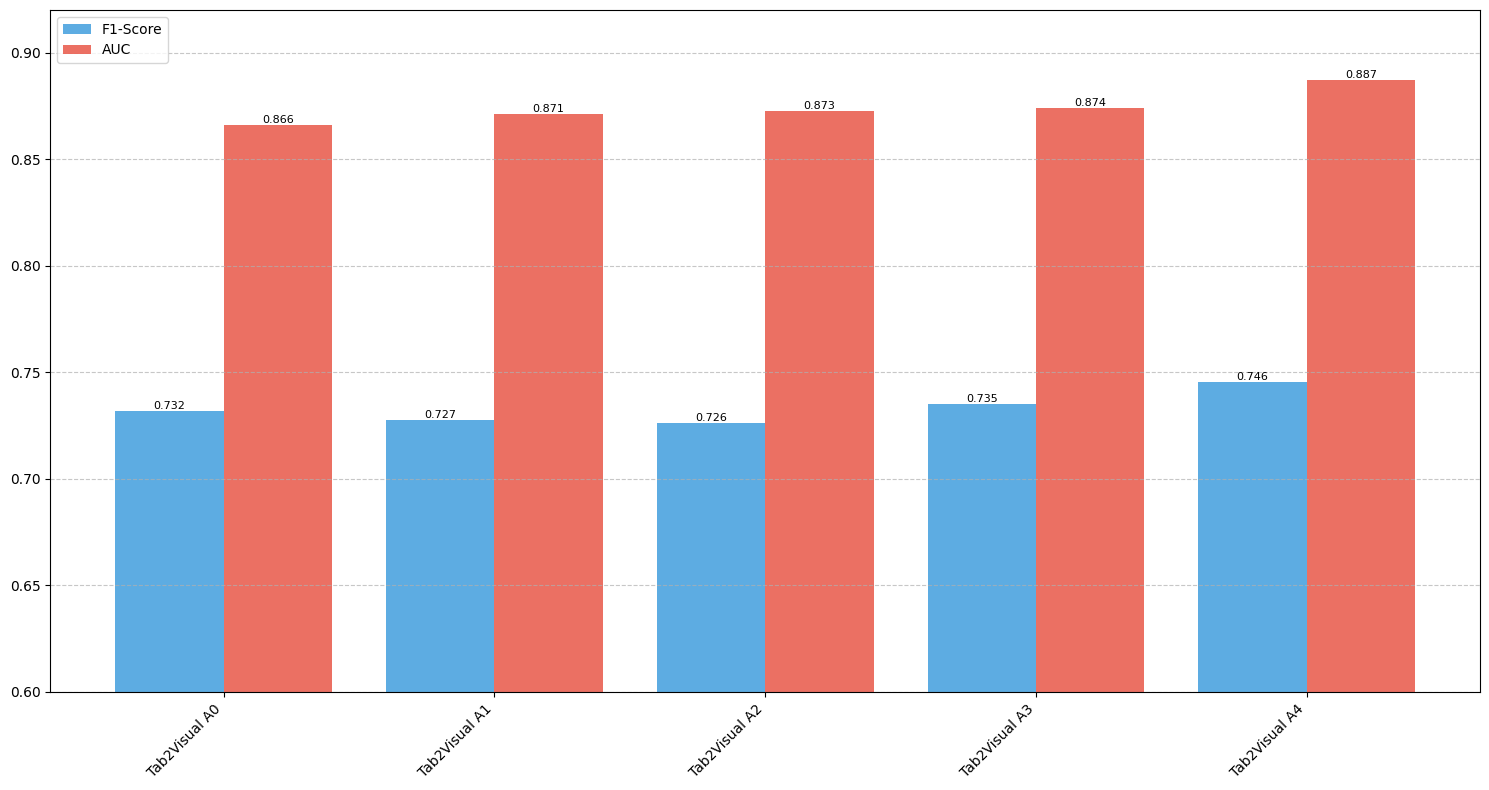}}
    \caption{Impact of augmentation levels (from \( \mathcal{A}_0 \) to \( \mathcal{A}_4 \)) on Tab2Visual's average performance on smaller datasets. Performance is measured in terms of F1-score and AUC. Higher values indicate better performance.}
    \label{fig:aug_effect}
\end{figure}

Tab2Visual incorporates image augmentation techniques to enhance data diversity and effectively increase the dataset size. Figure~\ref{fig:aug_effect} illustrates the impact of varying augmentation levels from $\mathcal{A}_0$ to $\mathcal{A}_4$ on Tab2Visual's average performance across smaller datasets. A consistent improvement is observed in both average F1-score and AUC with increasing augmentation levels. For instance, AUC accuracy increases from 86.6\% at $\mathcal{A}_0$ to 88.7\% at $\mathcal{A}_4$, representing a 2.1\% gain when each training sample is augmented four times using the methods outlined in Algorithm~\ref{alg:alg2}. Similarly, an approximately 1.4\% gain in F1-score is observed at augmentation level $\mathcal{A}_4$. These results unequivocally demonstrate that augmentation significantly enhances Tab2Visual's generalization ability and improves its predictive accuracy.

%Tab2Visual's image augmentation techniques are aimed to enhance data diversity and size.
%Figure~\ref{fig:aug_effect} illustrates the impact of image augmentation on Tab2Visual's average performance across smaller datasets. As the augmentation level increases from \(\mathcal{A}_0\) to \(\mathcal{A}_4\), we observe a consistent improvement in both average F1-score and AUC. For example, the accuracy in terms of AUC has increased from 86.6\% on \(\mathcal{A}_0\) to 88.7\% on \(\mathcal{A}_4\). That is, the accuracy gain is about 2.1\% when each training sample is augmented 4 times using the proposed augmentation methods of Algorithm~\ref{alg:alg2}. Similarly, the gain in terms of F1-score is about 1.4\% at $\mathbf{K}=4$. These results clearly demonstrate that augmentation significantly enhances Tab2Visual's generalization ability and improves its predictive accuracy.
%

\subsection{Training from Scratch} 
We here investigate the benefits of transfer learning by comparing the performance of EfficientNetV2-B0 trained from scratch (without ImageNet pre-training) against the pre-trained model.  All other experimental settings are matched to those described in Section~\ref{results_config}. Experiments are performed on the original, unaugmented datasets.  The results in Table~\ref{table:scratch_vs_transfer} clearly demonstrate the advantages of transfer learning.  On smaller datasets, pre-training yields substantial improvements, with average AUC gains of 7.5\% (peaking at 11\% for the glass dataset) and average F1-score gains of 6.7\%.  Smaller improvements (around 1\% for both AUC and F1-score) are also observed on larger datasets. These results suggest that the abundance of data can reduce the necessity for extensive pre-training.

%To assess the impact of transfer learning, we conduct an experiment where the EfficientNetV2-B0 model was trained from scratch without any pre-training on ImageNet. All other experimental parameters remained consistent with the previous experiments (Section~\ref{results_config}). Experiments are carried out on the original datasets $\mathcal{A}_0$, without any augmentation. The results, presented in Table~\ref{table:scratch_vs_transfer}, demonstrate the significant advantage of leveraging transfer learning. Notably, on smaller datasets, pre-training the EfficientNetV2-B0 model leads to substantial improvements in F1-score and AUC. On average, the AUC accuracy gain for all smaller datasets is about 7.5\%, reaching a high gain of 11\% on the glass dataset. The average F1-score gain across smaller datasets is 6.7\%. These highlight the crucial role of transfer learning in enhancing the model's performance when data is limited. While transfer learning also improves performance on larger datasets, the effect is less pronounced. Both AUC and F1-score gains, averaged on the larger datasets, are about 1\%. This suggests that the availability of ample data can partially mitigate the need for extensive pre-training.
%
%

\begin{table}[H]
    \caption{Training Tab2Visual from Scratch vs. Transfer Learning on Different Datasets.}
    \label{table:scratch_vs_transfer}
    \scriptsize
    \begin{tabularx}{\textwidth}{C C C C C}
    \toprule
    \textbf{Dataset} &
    \multicolumn{2}{c}{\textbf{From Scratch Learning}} &
    \multicolumn{2}{c}{\textbf{Transfer Learning}} \\
    \cmidrule(lr){2-3} \cmidrule(lr){4-5}
    & \textbf{F1-Score} & \textbf{AUC} & \textbf{F1-Score} & \textbf{AUC} \\
    \midrule
    \textbf{HRT}        & 0.4439 & 0.6933 & 0.5246 & 0.7346 \\
    \midrule
    \textbf{JU}         & 0.6690 & 0.8016 & 0.7510 & 0.8672 \\
    \midrule
    \textbf{DIA}        & 0.5484 & 0.6889 & 0.6262 & 0.7790 \\
    \midrule
    \textbf{BC}     & 0.8825 & 0.8984 & 0.9561 & 0.9884 \\
    \midrule
    \textbf{GL}         & 0.5474 & 0.7504 & 0.6014 & 0.8604 \\
    \midrule
    \textbf{PE}   & 0.8983 & 0.9141 & 0.9331 & 0.9655 \\
    \midrule
    \textbf{SAT}        & 0.8672 & 0.9661 & 0.8761 & 0.9826 \\
    \midrule
    \textbf{EMP}        & 0.9389 & 0.9751 & 0.9353 & 0.9742 \\
    \midrule
    \textbf{EG}    & 0.9291 & 0.9567 & 0.9513 & 0.9830 \\
    \midrule
    \textbf{TEL}        & 0.7664 & 0.9053 & 0.7823 & 0.9100 \\
    \bottomrule
    \end{tabularx}
\end{table}

%%%%%%%%%%%%%%%%%%%%%%%%%%%%%%%%%%%%%%%%%%%%%%%%%%%%%%%%%%%%
\subsection{CNNs vs ViTs}
The proposed Tab2Visual methodology is general and can be used with several deep network architectures as a backbone.
In our earlier experiments, we have demonstrated its performance using a CNN backbone from the EfficientNet family~\cite{tan2019efficientnet}.  
To further investigate the impact of the backbone model, in another series of experiments, we replace EfficientNetV2-B0 with EfficientViT~\cite{liu2023efficientvit}, a memory-efficient ViT optimized for both accuracy and efficiency. EfficientViT leverages Cascaded Group Attention (CGA) to significantly reduce computational overhead while effectively capturing local and global image features.
This enables us to directly compare the performance of ViTs, which excel at capturing long-range dependencies through attention mechanisms, against CNNs like EfficientNetV2 within the Tab2Visual framework.

We compare the results of EfficientViT against those obtained using EfficientNetV2 as the backbone model. We utilize EfficientViT pre-trained on ImageNet-1k~\cite{rw2019timm} and fine-tune it on each dataset, following the experimental setup described in Section~\ref{results_config}. The performance of EfficientViT is then compared with that of EfficientNetV2 in terms of average AUC and F1-score on smaller datasets for various augmentation levels as well as on larger datasets without augmentation, see Table~\ref{table:vit_cnn}.

\begin{table}[H]
  \caption{Comparison of EfficientNetV2 and EfficientViT Backbones for Tab2Visual: Average Performance on Various Datasets.}
  \label{table:vit_cnn}
  \centering
  \scriptsize  % <-- You can switch to \footnotesize, \tiny, etc.
  \begin{tabularx}{\textwidth}{C C C C C C}
      \toprule
      \multicolumn{2}{c}{\textbf{Dataset}} &
      \multicolumn{2}{c}{\textbf{Tab2Visual with CNN}} &
      \multicolumn{2}{c}{\textbf{Tab2Visual with ViT}} \\
      \cmidrule(lr){1-2} \cmidrule(lr){3-4} \cmidrule(lr){5-6}
      \textbf{Type} & \textbf{Augmentation} & \textbf{F1-score} & \textbf{AUC} & \textbf{F1-score} & \textbf{AUC} \\
      \midrule
      \multirow{6}{*}{\textbf{Smaller Datasets}} & \( \mathcal{A}_0 \) & 0.73 & 0.87 & 0.71 & 0.84 \\
      & \( \mathcal{A}_1 \) & 0.73 & 0.87 & 0.71 & 0.85 \\ 
      & \( \mathcal{A}_2 \) & 0.73 & 0.87 & 0.72 & 0.85 \\ 
      & \( \mathcal{A}_3 \) & 0.74 & 0.87 & 0.72 & 0.85 \\ 
      & \( \mathcal{A}_4 \) & 0.75 & 0.89 & 0.73 & 0.87 \\ 
      \midrule
      \textbf{Larger~Datasets} & \( \mathcal{A}_0 \) & 0.89 & 0.96 & 0.90 & 0.97 \\ 
      \bottomrule
  \end{tabularx}
\end{table}

EfficientNetV2 demonstrates superior performance on smaller datasets ($\leq$ 1000 samples) compared to EfficientViT, achieving higher F1-scores (0.73-0.75 vs. 0.71-0.73) and AUC values (0.87-0.89 vs. 0.84-0.87).
Interestingly, the performance trend reverses on larger datasets ($\ge$ 6000 samples), with EfficientViT achieving slightly higher F1-scores (0.90 vs. 0.89) and AUC values (0.97 vs. 0.96) than EfficientNetV2. Consistent with our previous findings in Section~\ref{sec:augmentation}, increasing data augmentation continues to enhance the performance of both models across all dataset sizes.

Our findings suggest that EfficientNetV2's powerful feature extraction capabilities make it a better choice for smaller datasets within our Tab2Visual framework.  However, on larger datasets, EfficientViT offers a competitive advantage, likely owing to its ability to effectively model long-range dependencies and complex feature interactions through its attention mechanisms.  EfficientViT's optimized architecture, incorporating Cascaded Group Attention~\cite{liu2023efficientvit}, enables it to scale effectively and maintain high accuracy with increased data.

%Table~\ref{table:vit_cnn} reveals that EfficientNetV2 consistently outperforms EfficientViT on smaller datasets in terms of both F1-score (0.73-0.75) and AUC (0.87-0.89). While EfficientViT achieves comparable results, with F1-scores ranging from 0.71 to 0.73 and AUC values between 0.84 and 0.87, it slightly lags behind EfficientNetV2 on these smaller datasets. 
%
%Interestingly, the performance trend shifts on larger datasets ($\ge$ 6000 samples), where EfficientViT shows a slight advantage over EfficientNetV2 in both F1-score (0.90 vs. 0.89) and AUC (0.97 vs. 0.96).  This suggests that ViT architectures, such as EfficientViT, are well-suited for leveraging larger datasets.  Across both smaller and larger datasets, the results here re-confirm that increasing data augmentation using the proposed set of operations consistently improves the performance of both EfficientNetV2 and EfficientViT.
%
%These findings suggest that EfficientNetV2, with its strong feature extraction capabilities, may be better suited for handling smaller datasets in our Tab2Visual approach. However, EfficientViT demonstrates a competitive advantage on larger datasets, likely due to its ability to effectively capture long-range dependencies and process complex feature interactions through its attention mechanisms. EfficientViT's optimized architecture, incorporating Cascaded Group Attention (CGA) for improved memory and computational efficiency, allows it to scale effectively and maintain high accuracy with larger datasets.
%
%%%%%%%%%%%%%%%%%%%%%%%%%%%%%%%%%%%%%%%%%%%%%%%%%%%%%%%%%%%%%%%
\subsection{Different Feature Arrangements}
\begin{figure}
    \centerline{\includegraphics[width=\linewidth]{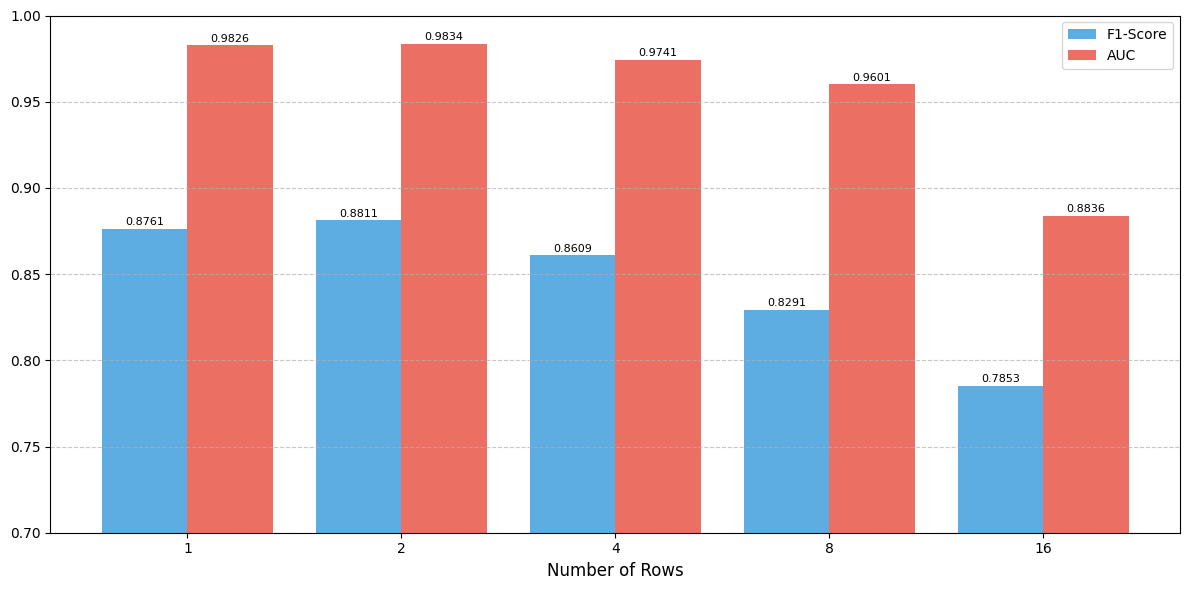}}
    \caption{Impact of feature arrangement on Tab2Visual's performance: F1-score and AUC for different row configurations on the Satellite Dataset.}
    %Summary results of Tab2Visual performance on Satellite dataset when encoding the samples with different number of rows.}
    \label{fig:feat_arrnge}
\end{figure}

We then explore how feature arrangement within the Tab2Visual image representation affects performance.  Using the Satellite dataset, which has the largest number of features (37) among our datasets, we vary the number of rows ($r$ in Algorithm~\ref{alg:alg1}) used to arrange the features (refer to Fig.~\ref{fig:bar_40_combined}).  EfficientNetV2 is fine-tuned for each arrangement, following the experimental setup in Sections~\ref{tab2vis_config} and~\ref{results_config}. Figure~\ref{fig:feat_arrnge} presents the results.

%
%We then investigate the impact of feature arrangement in the image presentations of the tabular data. We conduct experiments on the Satellite dataset, which has the largest number of features (37) among our datasets. We vary the number of rows ($r$ in Algorithm~\ref{alg:alg1}) used to arrange the features in the Tab2Visual representation, as described in Section~\ref{sec:t2v} and illustrated in Fig.~\ref{fig:bar_40_combined}. Using the experimental setup outlined in Sections~\ref{tab2vis_config} and~\ref{results_config}, we fine-tune EfficientNetV2 for each feature arrangement and evaluate its performance. The results of this analysis are presented in Fig.~\ref{fig:feat_arrnge}.
%

Our results in Fig.~\ref{fig:feat_arrnge} show a strong influence of feature arrangement (number of rows) on EfficientNetV2's performance on the Satellite dataset.  We observe optimal performance with 1 or 2 rows, with negligible differences between these two.  Increasing the number of rows results in a gradual decline in both F1-score and AUC, with approximately 10\% drop in both metrics at $r=16$.  This suggests that compact arrangements tend to be more effective for preserving feature relationships and maximizing generalization in Tab2Visual.  Therefore, careful feature arrangement is essential for optimizing model performance on a given dataset.

%
%
%Fig.~\ref{fig:feat_arrnge} demonstrates that the number of rows used to arrange features significantly influences EfficientNetV2's performance on the Satellite dataset. Optimal performance is observed with 1 or 2 rows, with minimal performance differences between these configurations. As the number of rows increases, a gradual decline in F1-score and AUC is observed. This suggests that compact feature arrangements with fewer rows are more effective in preserving feature relationships and maximizing model generalization within the Tab2Visual framework. This outcome highlights the importance of carefully considering feature arrangement in the Tab2Visual framework to optimize model performance on a given dataset.

%%%%%%%%%%%%%%%%%%%%%%%%%%%%%%%%%%%%%%%%%%%%%%%%%%%%%%%%%%%%%%%%%%%%%%%%%%%%%%%%%%%
\subsection{Time Performance} 
\begin{figure}
    \centerline{\includegraphics[width=\linewidth]{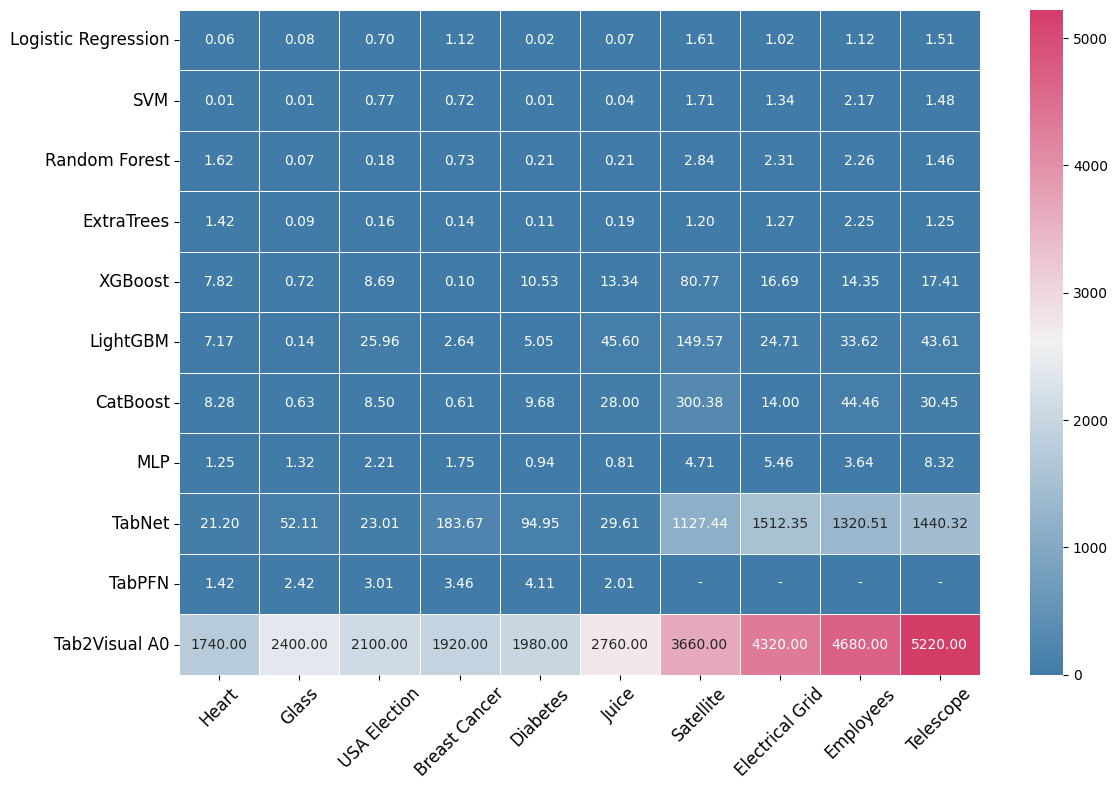}}
    \caption{Computational cost analysis: Heatmap of average 5-fold training times (in seconds) for different classifiers on various datasets.}
    \label{fig:tt_heatmap}
\end{figure}

Fig.~\ref{fig:tt_heatmap} illustrates the average 5-fold training time for each classifier across all datasets, measured on a workstation with the specifications outlined in Section~\ref{results_config}.
Traditional methods (Logistic Regression, SVM) exhibit the shortest training times, typically under 10 seconds, making them suitable for resource-constrained environments. Tree-based methods demonstrate moderate training times, generally completing within a minute for smaller datasets and under an hour for larger datasets. TabPFN stands out for its exceptional training speed on smaller datasets, consistently completing within 5 seconds. However, its applicability is limited to datasets smaller than 1000 samples~\cite{hollmann2022tabpfn}
\footnote{The recently released Version 2 extends its applicability to datasets with up to 10,000 samples~\cite{Hollmann_Nature2025}.}. 
TabNet exhibits longer training times, ranging from several minutes on smaller datasets to approximately 25 minutes on larger datasets, reflecting the complexity of its architecture. Tab2Visual, even without augmentation (i.e., $\mathcal{A}_0$), generally exhibits the longest training times, especially on larger datasets like Satellite and Electrical Grid. 
It is worth noting that Tab2Visual's training time includes the time required to convert tabular data into images.
These results underscore the usual trade-off between a model's complexity and its training time, highlighting the importance of selecting appropriate methods based on computational resources and task-specific requirements.
%These results highlight the trade-off between model complexity and training time, emphasizing the importance of selecting appropriate methods based on available computational resources and the specific requirements of the task.

During the test phase, all evaluated methods, including Tab2Visual, exhibit rapid inference times, typically completing within a fraction of a second (see Table~\ref{table:inference_times2}). Notably, the inference time for Tab2Visual includes the time required for the initial data-to-image transformation. 
These results demonstrate the potential of all evaluated methods, including Tab2Visual, for near real-time applications.

\begin{table}[H]
  \centering
  \scriptsize  % <-- Reduces the font size
  \caption{Average Inference Times for All Methods Across Various Datasets.\label{table:inference_times2}}
 
  % First tabular
  \begin{tabular}{l|c|c|c|c|c}
      \toprule
      Method & SVM & Logistic Regression & Extra Trees & Random Forest & LightGBM \\
      \midrule
      Inference Time (ms) & 0.2 & 1.2 & 1.0 & 1.7  & 1.5 \\
      \bottomrule
  \end{tabular}
  
  \vspace{2mm} % Adds a little vertical space before the second table
  
  % Second tabular
  \begin{tabular}{l|c|c|c|c|c|c}
      \toprule
      Method & XGBoost & CatBoost & MLP & TabNet & TabPFN & Tab2Visual A0 \\
      \midrule
      Inference Time (ms) & 2.0 & 2.4 & 10.0 & 7.0 & 101.0 & 289.0 \\
      \bottomrule
  \end{tabular}
  
  \normalsize % Revert to default font size if desired
\end{table}

%%%%%%%%%%%%%%%%%%%%%%%%%%%%%%%%%%%%%%%%%%%%%%%%%%%%%%%%%%%%%%%%%%%%%%%%%%%%%%%%%%%%%%%%%%%%%
\section{Conclusions}
\label{sec:conclusions}

This research addresses the critical challenge of limited data in tabular data classification, a prevalent issue exacerbated by the high proportion of small datasets observed in real-world applications. 
Deep learning models often struggle with such data, and traditional data augmentation and transfer learning methods are difficult to apply. 
To address this, we introduce Tab2Visual, a novel data transformation strategy that converts heterogeneous tabular data into
visual representations, unlocking the power of deep learning models, such as CNNs and ViTs.
Tab2Visual provides solutions for applying transfer learning and data augmentation to tabular data.
It leverages image augmentation to enhance model generalization and effectively increase training dataset size without additional data collection. We also introduce a new set of efficient and semantically meaningful image augmentation techniques tailored for Tab2Visual-generated images. Furthermore, 
Tab2Visual capitalizes on transfer learning, allowing for the fine-tuning of pre-trained models on specific tabular data classification tasks,
thereby reducing the reliance on large labeled datasets and allowing for knowledge sharing across different tabular data domains. Unlike existing tabular-to-image methods, Tab2Visual effectively handles data with limited features and offers tailored augmentation strategies, making it particularly well-suited for applications with small tabular datasets.

We have comprehensively evaluated our approach on ten diverse datasets, benchmarking its performance against a broad spectrum of machine learning algorithms, encompassing classical methods, tree-based ensembles, and state-of-the-art deep learning models specifically developed for handling tabular data.
 Our experimental results demonstrate Tab2Visual's effectiveness in classification problems with limited tabular data. On smaller datasets ($\leq$ 1000 samples), Tab2Visual has outperformed all compared methods, including specialized deep learning models like TabNet and TabPFN. Our experiments also demonstrate that tree-based ensembles exhibit rather better performances on larger datasets.

Furthermore, our study provides valuable insights into the key factors influencing Tab2Visual's performance:
\begin{itemize}
\item Impact of data augmentation: Our experiments demonstrate that augmentation significantly enhances Tab2Visual's generalization ability and improves its predictive accuracy, especially on smaller datasets.

\item Effectiveness of transfer learning: Transfer learning with pre-trained models significantly outperforms training from scratch, especially in data-limited scenarios.
%Compared to training from scratch, transfer learning with pre-trained models provides substantial performance improvements, particularly when data is limited.
%Leveraging pre-trained models, through transfer learning, has significantly enhanced Tab2Visual's performance, especially in data-limited scenarios, as compared to training a deep model from scratch.

\item Backbone model selection: Within the Tab2Visual framework, 
the CNN-based EfficientNetV2 model has outperformed  EfficientViT in our experiments on smaller tabular datasets ($\leq$ 1000 samples), while EfficientViT demonstrates superior performance on larger datasets. 

\item Influence of feature arrangement in the visual representations: Our empirical analysis shows that arranging the features in more compact representations tends to be more effective for maximizing generalization in Tab2Visual. 
\end{itemize}

% %%%%%%%%%%%%%%%%%%%%%%%%%%%%%%%%%%%%%%%%%%%%%%%%%%%%%%%
% %%% Move this to the Introduction in a good spot
% Tab2Visual presents a promising versatile and effective tool for tabular data classification, particularly in scenarios where data availability is limited. This study makes the following key contributions:
% %To summarize, our contributions are listed as follows:
% \begin{enumerate}
% \item We introduce a novel approach that transforms heterogeneous tabular data into visual representations for effective deep learning-based classification. Tab2Visual enables enhanced data augmentation and efficient transfer learning, reducing the reliance on large labeled datasets and allowing for knowledge sharing across different tabular data domains.

% \item We develop a new set of efficient and meaningful image augmentation techniques specifically designed for Tab2Visual-generated images, in order to increase data diversity and improve model generalization without requiring 
% additional data collection.

% \item We conduct a rigorous evaluation of Tab2Visual, comparing its accuracy and time performance against a wide range of machine learning algorithms across ten diverse datasets.

% \item We perform an in-depth analysis of key factors influencing Tab2Visual's performance, including the impact of augmentation, transfer learning, backbone model selection, and feature arrangement."
% \end{enumerate}
% %%%%%%%%%%%%%%%%%%%%%%%%%%%%%%%%%%%%%%%%%%%%%%%%%%%%%%%%%%%%%%%%%%%%%%%%%%%%

Several avenues for future research are currently being explored to further enhance Tab2Visual.
Firstly, we plan to investigate a broader range of image augmentation techniques beyond the set proposed in this work. Recognizing that a fixed augmentation policy may not be optimal for all datasets, we aim to adopt data-adaptive augmentation strategies~\cite{Hou_ICCV2023,Cheung_ICLR2022} to tailor augmentation policies specifically to each dataset, optimizing for improved generalization. 
Secondly, we will explore alternative feature arrangement strategies within the image representations, potentially by considering the similarity or correlation between features.
%leveraging feature similarity measures to enhance the informativeness of the visual representation. 
Thirdly, to further boost Tab2Visual's performance on larger datasets, we plan to investigate a wider range of deep learning backbones. 
Finally, we are actively working towards applying Tab2Visual to address real-world challenges, particularly in the domain of healthcare. We aim to utilize Tab2Visual to develop AI-driven prostate cancer diagnostic tools using limited tabular data comprising clinical biomarkers and symptoms variables.

%We are currently working on improving several aspects with Tab2Visual. 
%We plan to investigate other image augmentations in addition to the ones we have designed. 
%Since a fixed augmentation policy may not improve data diversity and model generalization for every dataset, 
%we plan to adopt data-adaptive policy~\cite{Hou_ICCV2023,Cheung_ICLR2022} for the images generated from a tabular dataset to produce a data-specific augmentation optimized for better generalization.
%Another direction for potential improvement is to study arranging the features in the image representations based
%in-between feature similarity. To boost Tab2Visual's performance on larger datasets, we plan to investigate several deep backbone models within the approach. Last but not least, we are are working to employ Tab2Visual to design prostate cancer diagnostics tools
%using limited tabular data consisting from clinical biomarkers and symptoms variables.

%%%%%%%%%%%%%%%%%%%%%%%%%%%%%%%%%%%%%%%%%%%%%%%%%%%%%%%%%%%%%%%%%%%%%%%%%
\section*{Acknowledgment}
This work has been funded in whole or in part with Federal funds from the National Cancer Institute, National Institutes of Health, under Task Order No. HHSN26110071 under Contract No. HHSN261201500003l; and also by NIH NIBIB P41EB015902 and NIH NIBIB P41EB028741. El-Melegy is supported through the Arab Fund Fellowship Program, Kuwait.
%%%%%%%%%%%%%%%%%%%%%%%%%%%%%%%%%%%%%%%%%%%%%%%%%%%%%%%%%%%%%%%%%%%%%%%%%%%%%%%%%%%%%%%%%%%%%%%5
\bibliographystyle{elsarticle-num} 
 \scriptsize
\bibliography{main.bib}

\end{document}